\documentclass[a4paper,fleqn]{cas-dc}

\usepackage[numbers]{natbib}

\usepackage{times}
\usepackage{epsfig}
\usepackage{graphicx}
\usepackage{amsmath}
\usepackage{amssymb}

\usepackage{bbm}
\usepackage{multirow}
\usepackage{algorithm}
\usepackage{algorithmic}
\usepackage{subfigure}
\usepackage{amsmath}
\usepackage{rotating}
\usepackage{gensymb}

\definecolor{mark}{RGB}{0,0,0}

\def\tsc#1{\csdef{#1}{\textsc{\lowercase{#1}}\xspace}}
\tsc{WGM}
\tsc{QE}
\tsc{EP}
\tsc{PMS}
\tsc{BEC}
\tsc{DE}

\begin{document}
\let\WriteBookmarks\relax
\def\floatpagepagefraction{1}
\def\textpagefraction{.001}
\shorttitle{Multi-Source Domain Adaptation for Panoramic Semantic Segmentation}
\shortauthors{J. Jiang et~al.}

\title [mode = title]{Multi-Source Domain Adaptation for Panoramic Semantic Segmentation}                      

\author[1]{Jing Jiang}
\affiliation[1]{organization={Faculty of Computing, Harbin Institute of Technology}, city={Harbin}, postcode={150001}, country={China}}

\author[2]{Sicheng Zhao}
\affiliation[2]{organization={BNRist, Tsinghua University},  city={Beijing}, postcode={100084}, country={China}}

\author[1]{Jiankun Zhu}

\author[3,4]{Wenbo Tang}
\affiliation[3]{organization={Didi Chuxing}, postcode={101300},  city={Beijing}, country={China}}
\affiliation[4]{organization={NavInfo}, postcode={100028}, city={Beijing}, country={China}}

\author[1]{Zhaopan Xu}

\author[1]{Jidong Yang}

\author[3]{Guoping Liu}

\author[3]{Tengfei Xing}

\author[3,4]{Pengfei Xu}


\author[1]{Hongxun Yao}

\nonumnote{E-mail address: 23s003045@stu.hit.edu.cn (J. Jiang).}

\begin{abstract}
Unsupervised \textcolor{mark}{domain adaptation methods for panoramic semantic segmentation utilize real pinhole images or low-cost synthetic panoramic images to transfer segmentation models to real panoramic images. However, these methods struggle to understand the panoramic structure using only real pinhole images and lack real-world scene perception with only synthetic panoramic images. Therefore, in this paper, we propose a new task, Multi-source Domain Adaptation for Panoramic Semantic Segmentation (MSDA4PASS), which leverages both real pinhole and synthetic panoramic images to improve segmentation on unlabeled real panoramic images. There are two key issues in the MSDA4PASS task: (1) distortion gaps between the pinhole and panoramic domains -- panoramic images exhibit global and local distortions absent in pinhole images; (2) texture gaps between the source and target domains -- scenes and styles differ across domains. To address these two issues, we propose a novel framework, Deformation Transform Aligner for Panoramic Semantic Segmentation (DTA4PASS), which converts all pinhole images in the source domains into distorted images and aligns the source distorted and panoramic images with the target panoramic images. Specifically, DTA4PASS consists of two main components: Unpaired Semantic Morphing (USM) and Distortion Gating Alignment (DGA). First, in USM, the Dual-view Discriminator (DvD) assists in training the diffeomorphic deformation network at the image and pixel level, enabling the effective deformation transformation of pinhole images without paired panoramic views, alleviating distortion gaps. Second, DGA assigns pinhole-like (pin-like) and panoramic-like (pan-like) features to each image by gating, and aligns these two features through uncertainty estimation, reducing texture gaps. DTA4PASS outperforms previous state-of-the-art methods by 1.92\% and 2.19\% in outdoor and indoor multi-source domain adaptation scenarios, respectively. The code is available at \href{https://github.com/jingjiang02/dta4pass}{https://github.com/jingjiang02/dta4pass}.
}
\end{abstract}

\maketitle
\section{Introduction}

\begin{figure}[t]
\centering
\includegraphics[width=0.98\linewidth]{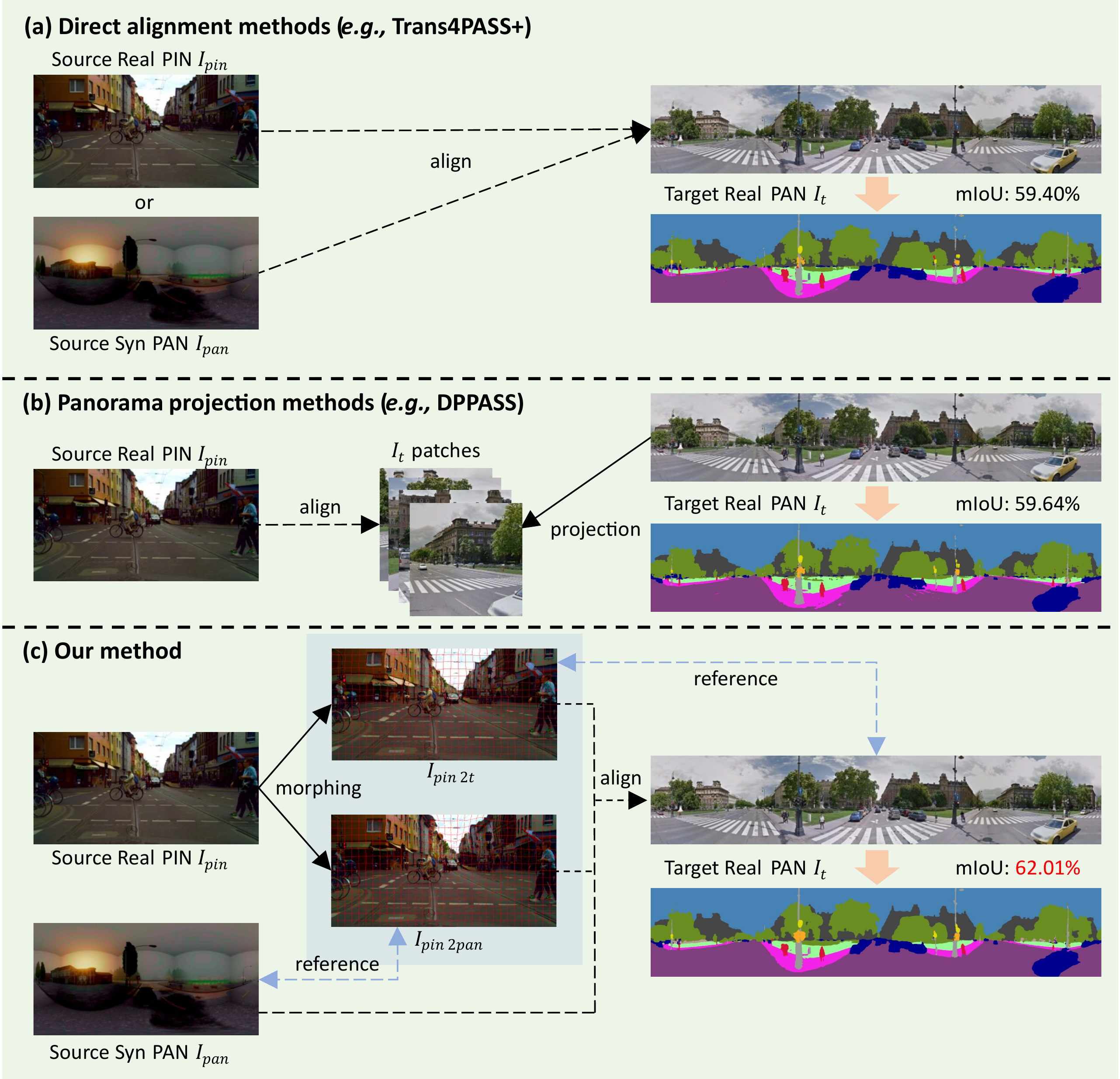}
\caption{Comparison of our method with other panoramic domain adaptation methods. The previous methods utilize only real pinhole or synthetic panoramic images. Meanwhile, they either (a) directly perform alignment, or (b) slice panoramic images into pinhole-like patches before alignment. (c) Our method converts all source pinhole images into distorted images and aligns source distorted images and source panoramic images with target panoramic images, perceiving both real-world scenes and panoramic structures.}
\label{fig:motivation}
\end{figure}

With the increasing demand for panoramic cameras, such as in autonomous driving, robot navigation, and video surveillance, it is particularly important to develop accurate and real-time semantic segmentation models for panoramic images. Recently, many segmentation methods for panoramic images have been proposed~\citep{orhan2022semantic,hu2022distortion,zhu2023research,li2023sgat4pass,yuan2023laformer,zheng2023complementary,yu2023panelnet,jin2024panoramic,shah2024multipanowise,guttikonda2024single,teng2024360bev,wei2024onebev,hu2024deformable,zheng2025open}.
Due to the higher cost of labeling panoramic images compared to pinhole images, there is still a lack of large-scale datasets for panoramic semantic segmentation.
Meanwhile, a large number of datasets for pinhole semantic segmentation have been accumulated~\citep{cordts2016cityscapes,zhou2017scene}, and many annotated synthetic panoramic images are generated at low cost through virtual engines~\citep{zheng2020structured3d, zhang2024behind}.  
Therefore, several domain adaptation (DA) methods have emerged that utilize these datasets to adapt the segmentation model to unlabeled real-world panoramic images~\citep{ma2021densepass,zhang2021transfer,zheng2023look,zhang2022bending,zhang2024behind,zheng2024semantics,zheng2024360sfuda++,kim2022pasts,zheng2023both,zhang2024goodsam,zhang2024goodsam++}. 

As described in Figure~\ref{fig:motivation} (a) and (b), these DA methods can be divided into two categories: direct alignment methods and panorama projection methods. The former directly aligns the features of the source and target images~\citep{ma2021densepass,zhang2021transfer,zhang2022bending,kim2022pasts,zheng2023look,zhang2024behind}, ignoring the distortion gap in appearance. The latter projects the target panoramic images into pinhole-like patches before aligning~\citep{zheng2023both,zhang2024goodsam,zhang2024goodsam++,zheng2024semantics,zheng2024360sfuda++}, which can destroy the panoramic structural information. 
Meanwhile, these methods can only utilize either real pinhole images or synthetic panoramic images. In fact, during adaptation, the segmentation model lacks an understanding of the panoramic structure when using only annotated real pinhole images, while it lacks the perception of real-world scenes when using only annotated synthetic panoramas.

Therefore, we propose a new task named Multi-source Domain Adaptation for Panoramic Semantic Segmentation (MSDA4PASS), which leverages both real pinhole and synthetic panoramic images to adapt the segmentation model to real panoramas. The MSDA4PASS task differs from traditional Multi-source Domain Adaptation (MSDA) tasks~\cite{sun2015survey,he2021multi,matsuzaki2023multi,zhao2021madan,pei2023multi,gao2023integrating,yu2023multi,zhang2022multi,nananukul2024multi,kang2023structure,gao2024multi,zhou2024simultaneous,xiao2024complex}, as it is a completely novel task with two points: (1) at least one of the source domains contains panoramic images; (2) the target domain consists entirely of panoramic images. This leads to two major challenges of the MSDA4PASS task: one is the distortion gap (\textit{e.g.}, object appearance distortion) between the pinhole and panoramic images, and the other is the texture gap (\textit{e.g.}, color, style) between the source and target images. 

To bridge the distortion and texture gaps between all source and target domains, we propose Deformation Transform Aligner for Panoramic Semantic Segmentation (DTA4PASS), demonstrated by Figure~\ref{fig:overall}. First, to mitigate the distortion gap, we propose Unpaired Semantic Morphing (USM), which transforms pinhole images into distorted images through an adversarial process. Second, to mitigate the texture gap, we propose Distortion Gating Alignment (DGA) to assign pin- and pan-like features to source distorted and source panoramic images with class-mixing and target panoramic images. DGA then performs alignments between these pin- and pan-like features through uncertainty estimation.

\textcolor{mark}{
Note that our method differs from existing domain adaptation methods in two key aspects: (1) For the pinhole domain adaptation methods and the direct alignment methods for panoramic images, our method considers the appearance distortion differences between pinhole and panoramic images that are ignored by these domain adaptation methods through the proposed USM image-to-image transformation. (2) For panorama projection methods, they project or crop the panoramic image into patches to alleviate the appearance distortion differences between pinhole and panoramic images. However, this can impair the segmentation network's global perception of the panoramic structure. Our method does not perform any visual processing on the panoramic images fed into the segmentation network. Instead, it converts the pinhole images into distorted images, which enhances the segmentation network's perception of distorted structures. Overall, our method is fundamentally different from existing domain adaptation methods, and the concept of converting a pinhole image into a distorted image is novel.
The key difference between the proposed DTA4PASS method and existing MSDA methods is that DTA4PASS addresses domain gaps between multiple pinhole and panoramic source domains and target domain at both the appearance level (image-to-image translation via USM) and the feature level (pinhole and panoramic feature alignment through DGA), which is an innovative approach.
}

We conduct experiments on two settings, namely outdoor scenarios: CS13~\citep{cordts2016cityscapes}, SP13~\citep{zhang2024behind}$\rightarrow$DP13~\citep{ma2021densepass}, and indoor scenarios: SPIN8~\citep{armeni2017joint}, S3D8~\citep{zheng2020structured3d}$\rightarrow$SPAN8~\citep{armeni2017joint}.
Extensive experimental results demonstrate the superiority of our method.
The contributions in this paper can be summarized as threefold:
\begin{itemize}
\item We introduce a new task, termed Multi-source
Domain Adaptation for Panoramic Semantic Segmentation (MSDA4PASS), to utilize both annotated pinhole images and low-cost synthetic panoramic images to adapt the segmentation model to real panoramic images.
\item We propose a novel Deformation
Transform Aligner for Panoramic Semantic Segmentation (DTA4PASS) framework for the MSDA4PASS task, consisting of USM and DGA. To bridge the distortion gap, the proposed USM morphs pinhole images into distorted images in an unsupervised adversarial manner. To bridge the texture gap, our DGA assigns pin- and pan-like features to multi-source mixed and target images at the pixel level, while performing alignment between them.
\item Extensive experiments in outdoor and indoor settings show that the proposed method outperforms the state-of-the-art methods by 1.92\% and 2.19\%, respectively.
\end{itemize}

The remaining part of this paper is organized as follows. Section~\ref{sec:related} introduces the works closely related to the MSDA4PASS task. Section~\ref{sec:methods} provides details of the MSDA4PASS task and the proposed DTA4PASS method. Section~\ref{sec:exps} presents the experimental setups, experimental results, ablation studies, and visualization analysis. In Section~\ref{sec:dis}, the advantages and limitations of the MSDA4PASS task and the proposed DTA4PASS method, as well as our future work, are further discussed. Finally, Section~\ref{sec:conclusion} concludes this paper.

\section{Related Works}
\label{sec:related}
\subsection{Panoramic Semantic Segmentation}

Panoramic semantic segmentation provides pixel-level semantic predictions for panoramic images with a $360^\circ$ field of view, and has important applications in both indoor and outdoor environments.
In outdoor scenarios, some specially designed deformable convolution networks~\citep{orhan2022semantic,hu2022distortion} are introduced to address distortions in panoramic images.
In indoor scenarios, attention mechanisms and deformable modules~\citep{zhu2023research,li2023sgat4pass,yuan2023laformer} are used to mitigate distortions. 
\citet{zheng2023complementary} combine horizontal and vertical data representations to better manage panoramic distortions. 
\citet{yu2023panelnet} use a panel-based representation of equirectangular projections to reduce panoramic distortion impacts. 
\citet{jin2024panoramic} enhance panoramic semantic segmentation by using HarDNet for efficient training, incorporating channel attention to highlight feature importance, and introducing a boundary loss to improve edge detection.
MultiPanoWise~\cite{shah2024multipanowise} introduces a holistic framework that employs a transformer-based multi-head architecture to jointly infer multiple pixel-wise and intrinsic signals from a single indoor panoramic image.
\citet{guttikonda2024single} propose a transformer-based cross-modal fusion architecture for multi-modal panoramic semantic segmentation, incorporating distortion-aware modules and cross-modal interactions to improve the integration of multi-modal features.

These methods rely on high-cost real panoramic segmentation annotations. In contrast, we utilize accumulated real pinhole images and low-cost synthetic panoramic images instead of annotated real panoramic images, enabling the segmentation model to perform effectively on unlabeled real panoramic images.

\subsection{Unsupervised Domain Adaptation}

Unsupervised domain adaptation (UDA) aims to transfer knowledge from labeled source domain samples to unlabeled target domain samples~\cite{tranheden2021dacs,huang2021domain,zhao2020review,hoyer2023mic,chen2023pipa,zhao2023toward,qayyum2024unsupervised,li2024multi,bi2024learning1,bi2024learning2,yi2024learning,bi2024learning3,bi2024generalized,zhao2024multi}.
Due to the lack of annotations for panoramic images, several UDA methods for panoramic images have been proposed.
To reduce the distortion gaps between the source and target domain, some methods utilize the attention mechanism~\citep{ma2021densepass, zhang2021transfer, zheng2023look}. 
Additionally, prototype alignment techniques~\citep{zhang2022bending,zheng2023look,zhang2024behind,zheng2024semantics,zheng2024360sfuda++} are adopted to reduce the domain gap in feature space. 
Trans4PASS~\cite{zhang2022bending} and its enhanced version, Trans4PASS+~\cite{zhang2024behind}, address domain adaptation for panoramic semantic segmentation by using deformable patch embedding and deformable MLP modules to handle distortions and deformations, along with mutual prototypical adaptation for cross-domain alignment.
PASTs~\cite{kim2022pasts} leverages a knowledge distillation approach with teacher-student branches to transfer segmentation knowledge from labeled pinhole images to panoramic images, effectively addressing structural distortions and enhancing segmentation performance.
DPPASS~\citep{zheng2023both} leverages equirectangular and tangent projections to address distortion gaps.
GoodSAM~\citep{zhang2024goodsam} and GoodSAM++~\cite{zhang2024goodsam++} introduce SAM~\citep{kirillov2023segment} to generate ensemble logits to achieve knowledge transfer from pinhole to panoramic images. 
360SFUDA~\cite{zheng2024semantics} and 360SFUDA++~\cite{zheng2024360sfuda++} address source-free unsupervised domain adaptation for pinhole-to-panoramic segmentation by employing tangent projection and panoramic prototype adaptation modules to adapt knowledge from pre-trained pinhole models to unlabeled panoramic images.

Naively applying existing panoramic UDA methods to the MSDA4PASS task can lead to suboptimal performance. They do not account for the distortion and texture gaps between the source domains. Meanwhile, previous pinhole multi-source UDA methods mainly focus on bridging the texture gaps between source and target domains~\citep{he2021multi,matsuzaki2023multi,zhao2019multi,zhao2021madan,pei2023multi,gao2023integrating,yu2023multi,liu2023pseudo,nananukul2024multi,kang2023structure} and cannot mitigate the distortion gaps between pinhole and panoramic images, resulting in unsatisfactory performance.

\subsection{Diffeomorphic Deformation Networks}

Diffeomorphic Deformation Networks are widely used in medical image registration tasks. Based on the given moving and fixed images, the diffeomorphic deformation network generates deformation fields, converting moving images into moved images, which are similar to fixed images. \citet{balakrishnan2019voxelmorph} propose a probabilistic generative model to generate deformation fields, enabling efficient alignment through either unsupervised or segmentation-assisted training. Subsequently, there emerged diffeomorphic deformation networks that utilize cycle consistency~\citep{kim2021cyclemorph}, Transformer~\cite{chen2022transmorph}, Swin Transformer~\cite{zhu2022swin} and Mamba~\cite{guo2024mambamorph} for stronger performance.
DaDA~\citep{jang2022dada} introduces distortion-aware domain adaptation for fisheye semantic segmentation, using relative distortion learning and affine transformations to address optical distortion shifts between source and target domains, enhancing adaptation in scenarios with unknown geometric deformations.

Our method differs from these approaches in several key ways. First, unlike deformation networks used in medical image registration, which rely on paired samples, we train our deformation network using unpaired pinhole and panoramic images due to the lack of paired original and deformed images.
Second, our method differs from DaDA in its focus on panoramic images within a multi-source DA setting, whereas DaDA is designed for fisheye images in a single-source DA setting. Further, the proposed Dual-view Discriminator (DvD) better captures both global and local distortion differences between pinhole and panoramic images, resulting in improved deformation fields.

\begin{table*}[h]
\caption{The main symbols used in this paper.}
\label{symbols}
\centering
\footnotesize
\begin{tabular}{llp{9cm}}
\hline
\textbf{Type} & \textbf{Symbols} & \textbf{Explanations} \\
\hline
\multicolumn{3}{l}{\textbf{Source pinhole domains}} \\
& $M$ & Number of source domains for pinhole images \\
& $U^i$ & The $i$-th source pinhole domain \\
& $|U^i|$ & Number of images of $U^i$ \\
& $I_{pin} = u^i_k$ & The $k$-th image in the $i$-th source pinhole domain $U^i$ \\
& $\epsilon^i_k$ & The label of source pinhole image $u^i_k$ \\
\hline
\multicolumn{3}{l}{\textbf{Source panoramic domains}} \\
& $N$ & Number of source domains for panoramic images \\
& $V^i$ & The $i$-th source panoramic domain \\
& $|V^i|$ & Number of images of $V^i$ \\
& $I_{pan} = v^i_k$ & The $k$-th image in the $i$-th source panoramic domain $V^i$ \\
& $\delta^i_k$ & The label of source panoramic image $v^i_k$ \\
\hline
\multicolumn{3}{l}{\textbf{Target domain}} \\
& $T$ & The target panoramic domain \\
& $|T|$ & Number of images of $T$ \\
& $I_{t} = t_k$ & The $k$-th image in the target panoramic domain $T$ \\
& $\hat{y}_k$ & The pseudo label of $t_k$ \\
\hline
\multicolumn{3}{l}{\textbf{Others}} \\
& $I_{pin2pan}$ & The deformed image of $I_{pin}$, referring to $I_{pan}$ \\
& $I_{pin2t}$ & The deformed image of $I_{pin}$, referring to $I_{t}$ \\
& $I_{a\&b}$ & The class-mixed image between $I_{a}$ and $I_{b}$ \\
& $x_i$ & Pinhole image $u^i_k$ from source pinhole domains \\
& $x_i^g$ & The grayscale image of $x_i$ \\
& $y_i$ & The label of $x_i$ \\
& $x_a$ & Panoramic images $v^i_k$, $t_k$ from source and target panoramic domains \\
& $x_a^g$ & The grayscale image of $x_a$ \\
& $\phi_{i2a}$ & The deformation field \\
& $\phi_{a2i}$ & The inverse deformation field of $\phi_{i2a}$ \\
& $x_{i2a}$ & The deformed image of $x_i$, referring to $x_{a}$ \\
& $y_{i2a}$ & The deformed label of $x_{i2a}$ \\
& $x_m$ & The class-mixed images between all source distorted images ($I_{pin2pan}$, $I_{pin2t}$, and $I_{pan}$) and target images $I_{t}$\\
& $y_m$ & The label of $x_m$ \\
\hline
\multicolumn{3}{l}{\textbf{Network}} \\
& $S$ & The semantic segmentation network \\
& $X^{tea}$ & The teacher network of $X$, which has the same structure as $X$ \\
& $f$ & The feature extractor of $S$ \\
& $g$ & The gating module of $S$ \\
& $h_{pin}$ & The segmentation head for the pinhole branch of $S$ \\
& $f_{pin}$ & The pin-like features obtained by $h_{pin}$ \\
& $h_{pan}$ & The segmentation head for the panoramic branch of $S$ \\
& $f_{pan}$ & The pan-like features obtained by $h_{pan}$ \\
& $h_{t}$ & The segmentation head for the target branch of $S$ \\
& $f_{t}$ & The gating fused features of $f_{pin}$ and $f_{pan}$ \\
& $F$ & The deformation network \\
& $D$ & The discriminators \\
& $D_e$ & The encoders of $D$, outputting image-level discrimination \\
& $D_d$ & The decoders of $D$, outputting pixel-level discrimination \\
\hline
\end{tabular}
\end{table*}

\begin{figure*}[!t]
\centering
\includegraphics[width=0.99\linewidth]{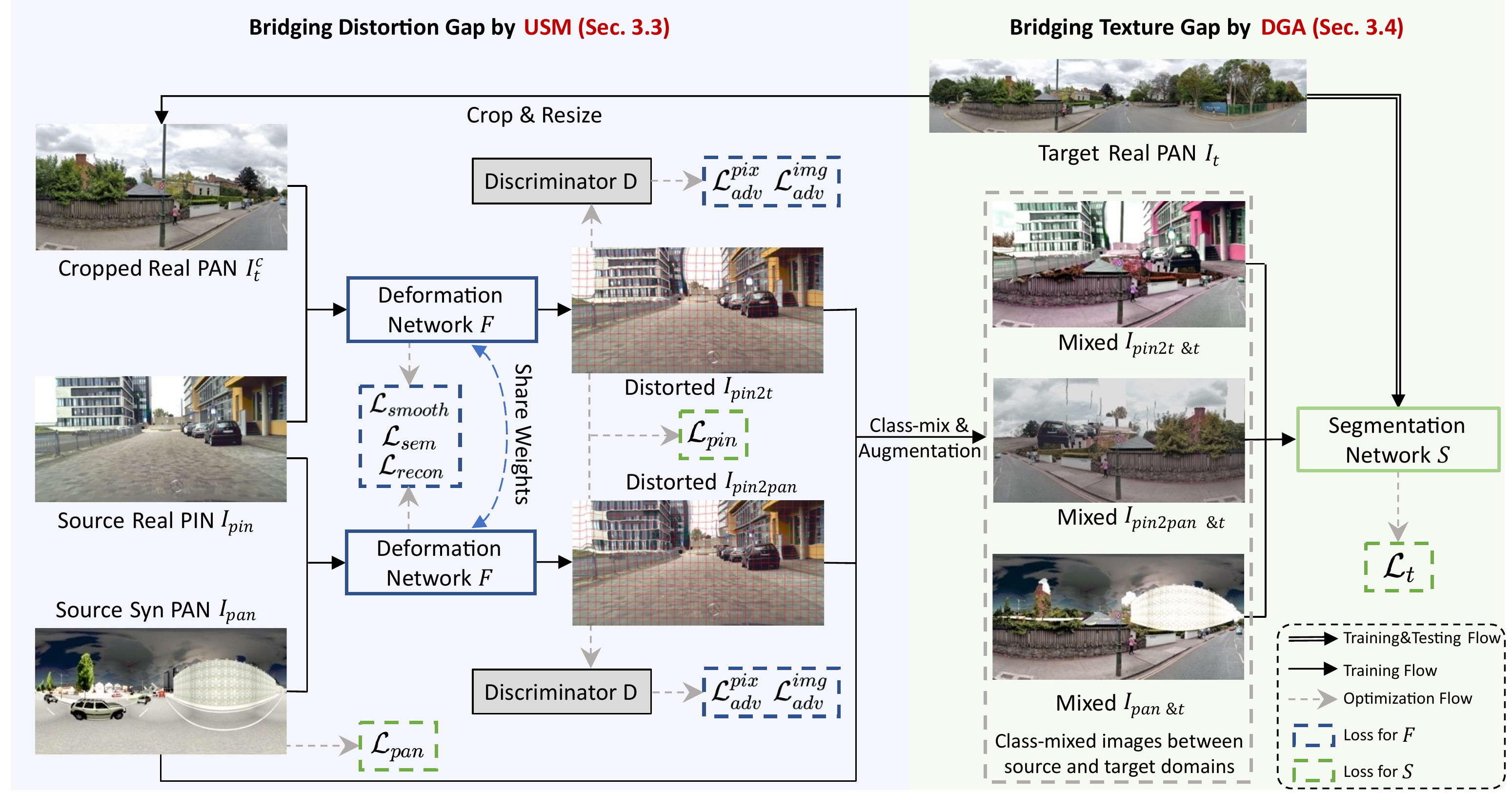}
\caption{An overall illustration of the proposed DTA4PASS. To bridge the distortion gap between pinhole and panoramic images, USM (Sec.~\ref{sec:USM}) converts all pinhole images into distorted images, referring to the source panoramic images $I_{pan}$ and the target panoramic images $I_{t}$. To bridge the texture gap between source and target images, DGA (Sec.~\ref{sec:DGA}) performs feature alignment between the class-mixed augmented source distorted/panoramic images and the target panoramic images.
}
\label{fig:overall}
\end{figure*}
\section{Methods}
\label{sec:methods}

\subsection{Preliminary}
We focus on the Multi-source Domain Adaptation for
Panoramic Semantic Segmentation (MSDA4PASS) task. Assuming we have $M$ source domains $\{U^i\}^{M}_{i=1}$ containing pinhole images $\{u^i_k\}^{M, |U^i|}_{i=1, k=1}$ with segmentation labels $\{\epsilon^i_k\}^{M, |U^i|}_{i=1, k=1}$, $N$ source domains $\{V^i\}^{N}_{i=1}$ including panoramic images $\{v^i_k\}^{N, |V^i|}_{i=1, k=1}$ with segmentation labels $\{\delta^i_k\}^{N, |V^i|}_{i=1, k=1}$, and target domain $T$ consisting of real panoramic images $\{t_k\}^{|T|}_{k=1}$. We aim to train the semantic segmentation network $S$, utilizing labeled source domains $\{U^i\}^{M}_{i=1}$ and $\{V^i\}^{N}_{i=1}$ to perform well on the unlabeled target domain $T$. Specifically, $S$ in this paper includes a feature extractor $f$, a gating module $g$ and three segmentation heads $h_{pin}, h_{pan},$ and $h_{t}$. Additionally, we have a teacher network $S^{tea}$ with the same structure as $S$, which is used for generating pseudo labels for unlabeled target domain samples $t_k$. The main symbols used in this paper are summarized in Table~\ref{symbols}.

\subsection{Model Overview}
DTA4PASS follows the paradigm of intermediate domain generation~\cite{zhao2024more}, first converting all source domain pinhole images into distorted images, and then aligning the source domain distorted pinhole images, source domain panoramic images, and target domain panoramic images. 
DTA4PASS mainly contains two parts: Unpaired Semantic Morphing (USM) and Distortion Gating Alignment (DGA). 
The former aims at bridging the distortion gap between pinhole and panoramic images, which converts the pinhole images of source domains into distorted images by utilizing unpaired pinhole and panoramic images in an unsupervised manner. Note that deformation transformation is only performed in pinhole images, as there is no distortion gap between panoramic images. The latter mainly focuses on bridging the texture gap between the converted multiple source domains and target domain, which assigns pin- and pan-like features to each image at the pixel-level, and aligns these two features. An illustration of DTA4PASS is shown in Figure~\ref{fig:overall}.

\subsection{Unpaired Semantic Morphing}
\label{sec:USM}

\begin{figure*}[!t]
\centering
\includegraphics[width=0.98\linewidth]{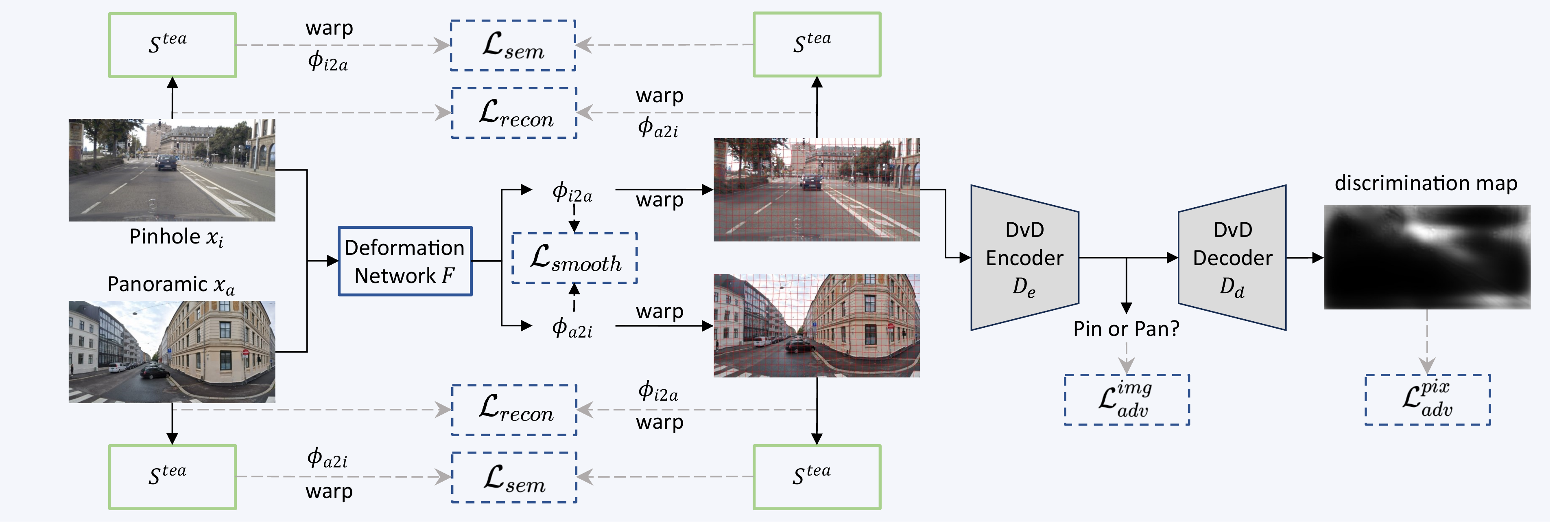}
\caption{Overview of Unpaired Semantic Morphing (USM), where brighter colors in pixel-level discrimination map indicate more pan-like. The pinhole and panoramic images $x_i$ and $x_a$ are fed into deformation network $F$ to obtain the deformation fields, as well as the deformed images $x_i \circ \phi_{i2a}$ and $x_a \circ \phi_{a2i}$. Afterward, the proposed Dual-view Discriminator (DvD) performs image-level discrimination $\mathcal{L}_{adv}^{img}$ and pixel-level discrimination $\mathcal{L}_{adv}^{pix}$ on the deformed images, assisting the deformation network $F$ in generating a deformation field that can transform the pinhole image $x_i$ into a distorted image similar to the panoramic image $x_a$, thereby mitigating the distortion gap between pinhole and panoramic domains.
}
\label{fig:architecture_morph}
\end{figure*}

\textbf{Motivation.} Directly aligning the source domain pinhole images, panoramic images, and target domain panoramic images is not appropriate due to significant distortion differences between the pinhole images in the source domains and the panoramic images. These distortion differences arise from the differing imaging principles between pinhole and panoramic images. Therefore, a direct and effective method is to introduce distortion into the pinhole images, ensuring that all images received by the segmentation network are distorted, thereby promoting its ability to perceive distortion.
As a result, we propose Unpaired Semantic Morphing (USM) to convert all source pinhole images into distorted images before conducting feature alignments, as shown in Figure~\ref{fig:architecture_morph}. USM consists of the deformation network $F$ and discriminators $D$, which learn distorted deformation fields in an adversarial manner using unpaired source pinhole images $x_i$ and panoramic images $x_a$ from the source and target domains. USM then applies the distorted deformation fields to the pinhole images $x_i$ and corresponding labels $y_i$ to obtain the deformed images $x_{i2a}$ and labels $y_{i2a}$.
It is worth noting that USM only deforms the source domain pinhole images, as there is no distortion gap between the source and target panoramas.

\textbf{Method.} Diffeomorphic deformation network $F$ is introduced to generate the deformation field $\phi_{i2a}$, making converted pinhole images $x_{i2a} = x_i \circ \phi_{i2a}$ as similar as possible to the panoramic images, where $x_i$ represents pinhole images.
Due to the absence of paired pinhole and panoramic images, unpaired images are used to train the deformation network $F$ in an adversarial manner.
Given the pinhole images $x_i \in \{u^j_k\}^{M, |U^j|}_{j=1, k=1}$ and the panoramic images $x_a \in \{v^j_k\}^{N, |V^j|}_{j=1, k=1} \cup \{t_k\}^{|T|}_{k=1}$, their grayscale images $x_i^g$, $x_a^g$ are obtained to eliminate the disturbance of color during the training of the deformation network $F$. Afterwards, $x_i^g$ and $x_a^g$ are fed into $F$ to obtain the deformation field $\phi_{i2a}$ and inverse field $\phi_{a2i}$:
\begin{equation}
    \phi_{a2i}, \phi_{i2a} = F(x_i^g, x_a^g).
\label{eq:gen_field}
\end{equation}

To make $x_i^g \circ \phi_{i2a}$ like $x_a^g$ in appearance, discriminators are introduced for adversarial learning. We propose Dual-view Discriminator (DvD) as the deformation discriminator $D$, which is a U-Net structure that can simultaneously output image-level and pixel-level predictions for pinhole/panoramic images, as described in Figure~\ref{fig:architecture_morph}. $x_i^g \circ \phi_{i2a}$ and $x_a^g$ are sent to $D$ to obtain image-level discriminator loss:
\begin{equation}
    \mathcal{L}_{dis}^{img}=-\mathbb{E}_{x_a^g}[\log D_{e}(x_a^g)] - \mathbb{E}_{x_i^g}[\log (1 - D_{e}(x_i^g \circ \phi_{i2a}))], 
\label{eq:loss_dis_img}
\end{equation} and pixel-level discriminator loss:
\begin{gather}\nonumber
    \mathcal{L}_{dis}^{pix}=-\mathbb{E}_{x_a^g}[\sum_{p,q} \log ([D(x_a^g)]_{p,q})] \\ - \mathbb{E}_{x_i^g}[\sum_{p,q} \log (1 - [D(x_i^g \circ \phi_{i2a})]_{p,q} )],
\label{eq:loss_dis_pix}
\end{gather}
where $D_{e}$ and $D_{d}$ are the encoder and decoder of $D = D_d \circ D_e$, outputting picture- and pixel-level predictions, respectively. $[D(x_a^g)]_{p,q}$ and $[D(x_i^g \circ \phi_{i2a})]_{p,q}$ are predictions at pixel $(p,q)$. The adversarial loss is formulated as: 
\begin{gather}
    \mathcal{L}_{adv}^{img}= - \mathbb{E}_{x_i^g}[\log (D_{e}(x_i^g \circ \phi_{i2a}))], \\
    \mathcal{L}_{adv}^{pix}= - \mathbb{E}_{x_i^g}[\sum_{p,q} \log ([D(x_i^g \circ \phi_{i2a})]_{p,q} )].
\end{gather}

In order to prevent excessive deformation, cycle consistent losses $\mathcal{L}_{recon}$ and $\mathcal{L}_{sem}$~\citep{jang2022dada} are introduced as follows:
\begin{gather} \nonumber
    \mathcal{L}_{recon}=\|x_i - (x_i \circ \phi_{i2a}) \circ \phi_{a2i}\|_1 \\ + \|x_a - (x_a \circ \phi_{a2i}) \circ \phi_{i2a}\|_1,
\end{gather}
\begin{gather} \nonumber
    \mathcal{L}_{sem}=\|S^{tea}(x_i \circ \phi_{i2a}) - S^{tea}(x_i) \circ \phi_{i2a}\|_1 \\ + \|S^{tea}(x_a \circ \phi_{a2i}) - S^{tea}(x_a) \circ \phi_{a2i}\|_1,
\end{gather}
where $||\cdot||_1$ is the L1 loss. Regularization loss $\mathcal{L}_{smooth}=smooth(\phi_{i2a})+smooth(\phi_{a2i})$ is also adopted to help the deformation network generate smooth deformation fields to avoid the introduction of noises. Here, $smooth$ is Kullback–Leibler divergence smooth loss~\citep{chen2022transmorph}.
Overall, the loss for USM is formulated as:
\begin{gather}\nonumber
    \mathcal{L}_{morph} = \frac{cur\_it}{max\_its} \cdot (\mathcal{L}_{recon} + \mathcal{L}_{sem}+\mathcal{L}_{smooth}) 
 \\ + \alpha \cdot (\mathcal{L}_{adv}^{img} + \mathcal{L}_{adv}^{pix}),
\label{eq:loss_morph}
\end{gather}
where $\alpha$ is the weight hyper-parameter, $cur\_it$ and $max\_its$ represent current and maximum iterations during training.

\subsection{Distortion Gating Alignment}
\label{sec:DGA}

\begin{figure*}[!t]
\centering
\includegraphics[width=0.98\linewidth]{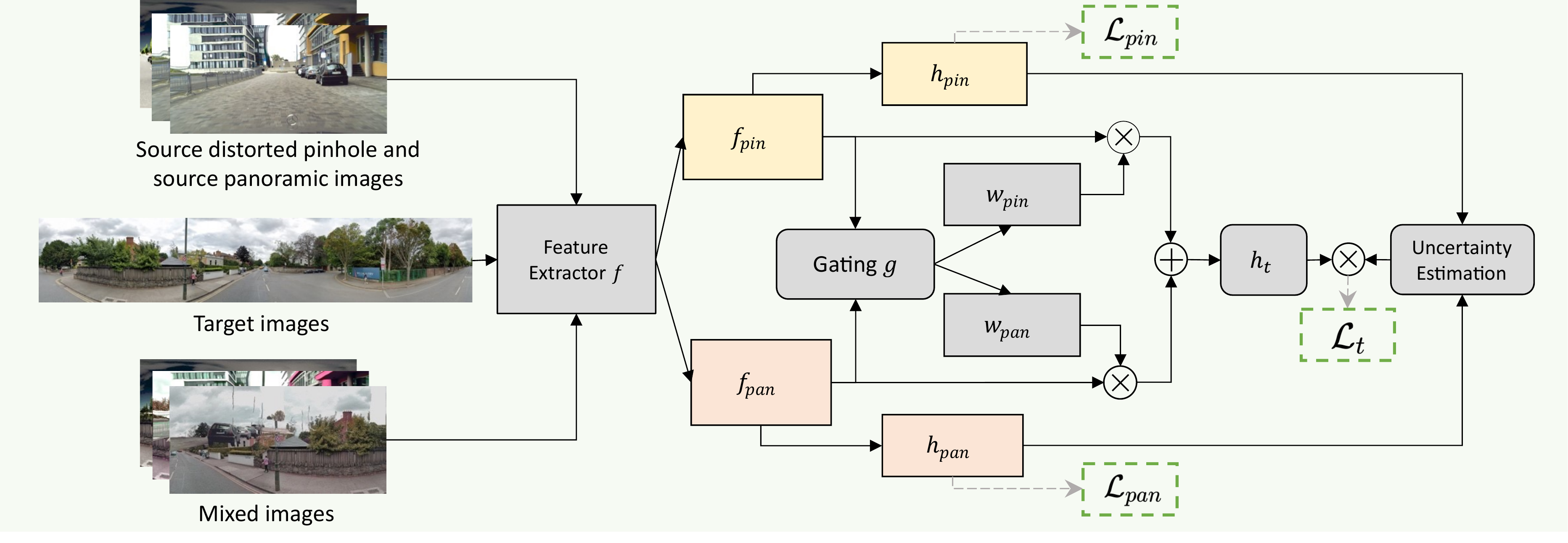}
\caption{Illustration of Distortion Gating Alignment (DGA). Given the source distorted pinhole images obtained from $\{U^i\}^{M}_{i=1}$ and the source panoramic images from $\{V^i\}^{N}_{i=1}$, they are fed into the pinhole (yellow) and panoramic (orange) branches respectively to train two auxiliary segmentation heads. For mixed and target images, they are fed into the target branch (grey). The gating module $g$ allocates pin-like features $f_{pin}$ and pan-like features $f_{pan}$ for input images at the pixel level. Finally, the uncertainty estimation module reduces the difference between these two features to alleviate the texture gap between source and target domains.}
\label{fig:architecture_seg}
\end{figure*}

\textbf{Motivation.} In Sec.~\ref{sec:USM}, USM is used to apply a deformation transformation to pinhole images $x_i$, converting them into distorted images $x_{i2a}$. At this point, all images in the source and target domains have distortion, and the distortion gap is alleviated. However, there are still texture differences between the source domains and the target domain, such as weather, style, scene, \textit{etc.}, which can lead to unsatisfactory generalization of the segmentation model. Therefore, aligning the source transformed pinhole images with the source panoramic images and the target panoramic images at the feature level is needed. Although the pinhole images are distorted, directly aligning the pinhole and panoramic images still complicates model optimization, particularly when multiple domains are involved.
Therefore, we propose Distortion Gating Alignment (DGA), which comprises a dual branch structure $h_{pin}$ and $h_{pan}$ to process the source domain pinhole and panoramic images separately to obtain pin-like and pan-like features to avoid conflicts, as shown in Figure~\ref{fig:architecture_seg}.
To achieve multi-source domain adaptation, the gating module $g$ is introduced to allocate pin-like features $f_{pin}$ and pan-like features $f_{pan}$ for source class-mixed images $x_m$ and target panoramic images $t_k$ at the pixel level. DGA then reduces the differences between pin-like and pan-like features through uncertainty estimation. The target branch $h_{t}$ of DGA ultimately uses gated fusion of pin-like and pan-like features to segment the target panoramic images, enabling the segmentation model to better adapt to distortion and generalize to the target panoramic domain.

\textbf{Method.} Considering the target domain sample $t_k$ has no labels, the pseudo label $\hat{y}_k$ of $t_k$ with a confidence threshold of $\eta$ is obtained through an Exponential Moving Average (EMA)~\citep{tarvainen2017mean} teacher:
\begin{equation}
    S^{tea} =  (h_{pin}^{tea} + h_{pan}^{tea} + h_{t}^{tea} \circ g^{tea}) \circ f^{tea},
\end{equation}
where $f^{tea}$ and $g^{tea}$ are the feature extractor and gating module, respectively. $h_{pin}^{tea}$, $h_{pan}^{tea}$, and $h_{t}^{tea}$ are pinhole, panoramic, and target branch segmentation heads, respectively. The weights $\theta^{tea}$ of  $S^{tea}$ are the exponential moving average of the weights $\theta$ of $S$ with smoothing factor $\gamma$:
\begin{equation}
    \theta^{tea}_{t+1} \leftarrow \gamma\theta^{tea}_{t} + (1-\gamma)\theta_{t},
\label{eq:update_tea}
\end{equation}
where $t$ denotes a training step.
Afterward, class-mix~\citep{tranheden2021dacs} is performed between the multiple source distorted images (\textit{i.e.}, $u^i_k \circ \phi_{i2a}$ and $v_k^i$) and the target domain panoramic images $t_k$ to obtain the mixed images $x_m$ and corresponding labels $y_m$.

For source domain distorted images from $\{u^i_k\}^{M, |U^i|}_{i=1, k=1}$ and $\{v^i_k\}^{N, |V^i|}_{i=1, k=1}$, they are fed into the pinhole and panoramic branches, respectively, to train two auxiliary segmentation heads $h_{pin}$ and $h_{pan}$ with standard Cross Entropy ($CE$) loss as follows:
\begin{gather}
    \mathcal{L}_{pin} = \sum^{M}_{i=1}\sum^{|U^i|}_{k=1}CE(h_{pin}(f(u^i_k \circ \phi_{i2a})), \epsilon^i_k \circ \phi_{i2a}), \\
    \mathcal{L}_{pan} = \sum^{N}_{i=1}\sum^{|V^i|}_{k=1}CE(h_{pan}(f(v^i_k)), \delta^i_k).
\label{eq:loss_pan}
\end{gather}
Afterwards, the source domains and the target domain are aligned through the target branch $h_{t}$. The mixed images $x_m$ and target images $t_k$ are fed into the feature extractor $f$ followed by $h_{pin}$ and $h_{pan}$ to obtain pin-like features $f_{pin}$ and pan-like features $f_{pan}$. Due to the fact that $h_{pin}$ and $h_{pan}$ mainly receive pinhole and panoramic images respectively, there exist domain gaps. To bridge the gap between them for multi-source domain alignment, a gating module $g$ is proposed to assign weights of $f_{pin}$ and $f_{pan}$ to each sample at pixel-level, as described below:
\begin{gather} \nonumber
    w = [w_{pin}, w_{pan}] = \sigma(g(concat(f_{pin}, f_{pan}))) \\
    f_{t} = \sum_{i \in \{pin, pan\}} w_{i} \odot f_{i},
\end{gather}
where $concat$ is concatenate operation, $\sigma$ is softmax operation, and $\odot$ is Hadamard product.
Meanwhile, uncertainty estimation~\citep{zheng2021rectifying} is introduced to perform alignment between the two features $f_{pin}$ and $f_{pan}$:
\begin{gather}
    D_{kl}=\mathbb{E}[KL(h_{pin}(f(x)), h_{pan}(f(x)))], \\
    \mathcal{L}_{t} = \sum_{x}\mathbb{E}[exp\{-D_{kl}\} \cdot CE(h_{t}(f_{t}), y)+D_{kl}],
\label{eq:loss_moe}
\end{gather}
where, $x \in \{x_m, t_k\}$, $y$ is the corresponding pseudo label of $x$, $KL$ is Kullback-Leibler Divergence, and $exp$ is an exponential operator.

In summary, the following loss function is obtained to train the segmentation network $S$:
\begin{equation}
    \mathcal{L}_{seg} = \mathcal{L}_{pin} + \mathcal{L}_{pan} + \beta\mathcal{L}_{t},
\label{eq:loss_seg}
\end{equation}
where $\beta$ is the weight hyper-parameter.

In the inference stage, the average fusion of three segmentation heads $h_{pin}, h_{pan}$, and $h_{t}$ is used as the final prediction result, as follows:
\begin{equation}
    S =  (h_{pin} + h_{pan} + h_{t} \circ g) \circ f.
\end{equation}

\subsection{Learning Procedure}

\begin{algorithm}[tb]
\caption{DTA4PASS Adaptation Procedure}
\label{alg:algorithm}
\textbf{Input}: source pinhole images $\{u^i_k\}^{M, |U^i|}_{i=1, k=1}$ with  labels $\{\epsilon^i_k\}^{M, |U^i|}_{i=1, k=1}$, source panoramic images $\{v^i_k\}^{N, |V^i|}_{i=1, k=1}$ with  labels $\{\delta^i_k\}^{N, |V^i|}_{i=1, k=1}$, target panoramic images $\{t_k\}^{|T|}_{k=1}$, diffeomorphic deformation network $F$, discriminators $D$, semantic segmentation network $S$, corresponding teacher network $S^{tea}$, and total iteration number $T_{total}$.\\
\begin{algorithmic}[1] 
\STATE Initialize $S$ and $S^{tea}$ with source pre-trained parameters, and initialize $F$ and $D$ randomly;
\FOR{iteration=1 to $T_{total}$}
\STATE Get source images $u^i_k$, $v^i_k$, labels $\epsilon^i_k, \delta^i_k$, and target images $t_k$;
\STATE Generate pseudo labels $\hat{y}_k$ for $t_k$ using $S^{tea}$;
\STATE Generate deformation fields $\phi_{i2a}$ and $\phi_{a2i}$ using Eq.~(\ref{eq:gen_field});
\STATE Warp images and labels, perform class-mix and augmentation between source deformed/panoramic images and target panoramic images;
\STATE Update $S$ using Eq.~(\ref{eq:loss_seg});
\STATE Update $F$ using Eq.~(\ref{eq:loss_morph});
\STATE Update $D$ using Eq.~(\ref{eq:loss_dis_img}) and Eq.~(\ref{eq:loss_dis_pix});
\STATE Update $S^{tea}$ using Eq.~(\ref{eq:update_tea});
\ENDFOR
\STATE \textbf{return} the adapted model $S$.
\end{algorithmic}
\label{al:overall}
\end{algorithm}

During the training of the segmentation network $S$, the deformation network $F$ is used to provide distorted images, and the segmentation teacher network $S^{tea}$ is adopted to obtain pseudo labels for the target domain panoramic images. During the training of the deformation network $F$, discriminators $D$ are used for adversarial learning, and the segmentation teacher network $S^{tea}$ provides the reconstruction loss.
The training of the proposed DTA4PASS involves alternating between the segmentation network, deformation network, and discriminators in an end-to-end manner to promote mutual improvement, as detailed in Algorithm~\ref{al:overall}.

During the inference stage, the segmentation teacher network $S^{tea}$, deformable network $F$, and discriminators $D$ are discarded, and only the segmentation network $S$ is used. Therefore, the efficiency of inference is not affected.

\section{Experiments}
\label{sec:exps}

\subsection{Experimental Settings}
\subsubsection{Datasets}

Extensive experiments are conducted on two scenarios.
One is the outdoor setting, composed of Cityscapes~\citep{cordts2016cityscapes} (CS13), SynPASS~\citep{zhang2024behind} (SP13), and DensePASS~\citep{ma2021densepass} (DP13); the other is the indoor setting, including Stanford2D3D~\citep{armeni2017joint} (SPIN8, SPAN8) and Structured3D~\citep{zheng2020structured3d} (S3D8).
Following \citet{zhang2024behind}, we adopt 13 shared classes under the outdoor setting: CS13,~SP13$\rightarrow$DP13 and 8 shared categories under the indoor setting: SPIN8,~S3D8$\rightarrow$SPAN8.

\textbf{Cityscapes}~\citep{cordts2016cityscapes} is a real pinhole dataset with a resolution of 1024$\times$2048, including 2,975, 500, and 1,525 images for training, validation, and testing respectively, and contains 19 classes for pinhole semantic segmentation.
\textbf{SynPASS}~\citep{zhang2024behind} has 5,700, 1,690, and 1,690 synthetic panoramic images for training, validation, and testing, respectively, consisting of 22 semantic categories, with a resolution of 1024$\times$2048. 
\textbf{DensePASS}~\citep{ma2021densepass} has 2,000 real panoramic images without labels for transfer learning, and 100 labeled panoramic images for testing, including 19 classes, with a resolution of 400$\times$2048.
\textbf{Stanford2D3D}~\citep{armeni2017joint} has 70,496 real pinhole images with a resolution of 1080$\times$1080 and 1,413 panoramic images with a resolution of 512$\times$1024, both containing 13 classes. \textbf{Structured3D}~\citep{zheng2020structured3d} is a synthetic dataset with 21,835 panoramic images and 29 categories, having a resolution of 512$\times$1024.

\subsubsection{Baselines}

We conduct comparisons with several methods in the following settings:

(1) \textbf{Source-only} includes the best result from training on a single source domain and testing in the target domain (referred to as Single-best), as well as the result of training on a combined source domain and testing in the target domain (referred to as Combined).

(2) \textbf{Single-best and Combined DA} refer to the best result of adapting from a single source domain to the target domain and the result of adapting from a combined source domain to the target domain, respectively. We compare our method with previous pinhole domain adaptation methods DACS~\citep{tranheden2021dacs},  MIC~\citep{hoyer2023mic}, and PiPa~\citep{chen2023pipa}, as well as panoramic domain adaptation methods DATR~\citep{zheng2023look} and Trans4PASS+~\citep{zhang2024behind}.
DACS~\citep{tranheden2021dacs} introduces a method for unsupervised domain adaptation in semantic segmentation by mixing images from source and target domains along with their corresponding labels and pseudo-labels to improve model generalization across domains. 
MIC~\citep{hoyer2023mic} enhances unsupervised domain adaptation by enforcing consistency between predictions of masked target images and their pseudo-labels, encouraging the model to learn spatial context relationships for improved recognition across domains.
PiPa~\citep{chen2023pipa} introduces a self-supervised learning framework for domain adaptive semantic segmentation that leverages pixel- and patch-wise correlations within images to enhance intra-domain feature learning and improve context consistency across different environments.
DATR~\citep{zheng2023look} introduces a distortion-aware attention mechanism and a class-wise feature aggregation module to efficiently address pixel distortion in panoramic images for unsupervised domain adaptation, optimizing feature similarity between domains without relying on geometric constraints.
Trans4PASS+~\citep{zhang2024behind} enhances panoramic semantic segmentation by incorporating deformable modules for handling distortions and deformations, alongside an improved mutual prototypical adaptation strategy for effective unsupervised domain adaptation in $360^\circ$ imagery.

(3) \textbf{Multi-source DA (MSDA)} includes pinhole methods LAB~\citep{he2021multi} and MS2PL~\citep{matsuzaki2023multi}.
LAB~\citep{he2021multi} introduces a collaborative learning framework for multi-source unsupervised domain adaptation in semantic segmentation, combining image translation for pixel alignment and ensemble-based pseudo-labeling to leverage unlabeled target data effectively.
MS2PL~\citep{matsuzaki2023multi} introduces a domain adaptive training approach for semantic segmentation that generates soft pseudo labels by weighting predictions from multiple source models based on domain similarity, using entropy-aware training to enhance robustness to misclassified pixels.

\subsubsection{Evaluation Metrics}
In this paper, we adopt the Mean Intersection over Union (mIoU) metric, which calculates the average overlap between predicted segmentation masks and ground truth labels, providing an intuitive measure of segmentation performance. In addition to the overall mIoU, we also employ category-level mIoU, which offers a per-class breakdown of mIoU scores, enabling a more granular analysis of model performance across various semantic categories.

\subsubsection{Implementation Details}

\textbf{Data Processing.}
\textcolor{mark}{The image resolution and random augmentation settings for training and testing of all the methods in the experiments are the same as those used in \citet{zhang2024behind}.} For model initialization, we train with only source domain samples and preprocess the source domain images to a resolution of 512$\times$512. For adaptation, the source domain and the cropped target domain images are processed to a resolution of 512$\times$1024. The target domain images are processed to resolutions of 400$\times$2048 and 1024$\times$2048, respectively, for outdoor and indoor scenes during both training and inference.
Following \citet{tranheden2021dacs}, we apply color jitter, Gaussian blur, and random flipping for augmenting the source class-mixed and target images. Additionally, the target images are augmented with random erasing. Further, we adopt the LAB color transformation~\citep{he2021multi} between the source and target images.

\textbf{Network Structures.}
\textcolor{mark}{For a fair comparison, Segformer-B2~\citep{xie2021segformer} is used as the feature extractor $f$ for all the methods in the experiments.} DMLPv2~\citep{zhang2024behind} is adopted as the pinhole and panoramic segmentation heads $h_{pin}$ and $h_{pan}$. For the target segmentation head $h_{t}$, a single-layer lightweight convolutional network is used.
For the gating module $g$, we use a convolutional layer with a 3$\times$3 kernel and a stride of 1, with 128 channels, followed by a BatchNorm layer and ReLU activation. Finally, a convolutional layer with a 3$\times$3 kernel and a stride of 1, with 2 channels, is adopted to obtain pixel-level pinhole and panoramic feature weights.

TransMorph~\cite{chen2022transmorph} is used as the diffeomorphic deformation network $F$.
The network architecture of the proposed discriminator DvD is based on the discriminator from StyleGAN2~\cite{karras2020analyzing}. We extend the original discriminator $D$ to a U-Net architecture, utilizing the features from the middle layer (with the lowest resolution) for image-level discrimination and the features from the final layer (with the highest resolution) for pixel-level discrimination. For any pair of pinhole and panoramic domains (both source and target), we share the deformable network $F$ to enhance efficiency. However, each pair of pinhole and panoramic domains has an independent discriminator $D$ to prevent insufficient discrimination and unsatisfactory deformation. Specifically, in this paper, we employ two discriminators: one to distinguish between the source pinhole and the source panoramic domain, and the other to distinguish between the source pinhole and the target panoramic domain.

\begin{table*}[!t]
  \caption{Comparison with the state-of-the-art DA methods on outdoor scenarios. * indicates the results provided in other papers~\citep{zheng2023look,zhang2024behind}.}
  \label{outdoor_table}
  \centering
  \setlength{\tabcolsep}{2pt} 
  \small
  \resizebox{\linewidth}{!}{%
  \begin{tabular}{c|c|c|lllllllllllll}
    \toprule
Standards&Methods&mIoU&Road&S.walk&Build.&Wall&Fence&Pole&Tr.light&Tr.sign&Veget.&Terrain&Sky&Person&Car\\
    \midrule
    \multirow{2}{*}{Source-only}& Single-best*  &51.48 &76.45 &40.52 	&86.16 	&28.70 	&43.77 	&26.93 	&15.75 	&16.71 	&79.91 	&32.48 	&93.76 	&49.21 	&78.87\\
    ~& Combined &52.39 &79.25 	&51.23 	&85.63 	&26.48 	&39.45 	&27.39 	&21.75 	&13.46 	&77.54 	&40.24 	&92.80 	&49.72 	&76.21\\
    \midrule
    \multirow{5}{*}{Single-best DA}& DACS & 53.10 &80.72 	&51.03 	&85.19 	&27.56 	&40.11 	&27.59 	&10.82 	&18.35 	&78.27 	&44.47 	&92.96 	&53.12 	&80.04\\
    ~& MIC &52.55 &79.89 	&45.98 	&85.91 	&32.39 	&36.78 	&29.32 	&15.91 	&15.47 	&77.54 	&44.06 	&92.83 	&49.66 	&77.45\\
    ~& PiPa &53.29 &80.28 	&47.79 	&86.24 	&30.08 	&41.96 	&28.49 	&12.68 	&16.98 	&78.51 	&41.99 	&93.74 	&55.21 	&78.88\\
    ~& DATR* &54.05 &79.07 	&52.28 	&85.98 	&33.38 	&45.02 	&\textbf{34.47} 	&26.15 	&18.27 	&78.21 	&26.99 	&94.02 	&51.21 	&77.62\\
    ~& Trans4PASS+* &55.24 &\textbf{82.25} 	&\textbf{54.74} 	&85.80 	&31.55 	&47.24 	&31.44 	&21.95 	&17.45 	&79.05 	&45.07 	&93.42 	&50.12 	&78.04\\
    \midrule
    \multirow{4}{*}{Combined DA}&DACS &52.75 &79.55 	&51.13 	&86.43 	&25.17 	&39.50 	&28.31 	&24.97 	&14.96 	&77.73 	&34.81 	&92.39 	&51.38 	&79.48\\
    ~ & MIC &53.03 &78.47 	&52.80 	&86.27 	&29.68 	&39.19 	&27.28 	&24.17 	&14.23 	&73.90 	&43.67 	&89.87 	&51.10 	&78.73\\
    ~ & PiPa &53.54 &80.14 	&51.51 	&86.38 	&27.36 	&41.83 	&29.98 	&\textbf{27.99} 	&14.53 	&77.98 	&34.88 	&92.87 	&51.67 	&78.93\\
    ~ & Trans4PASS+ &54.50 &80.95 	&54.11 	&86.70 	&31.04 	&42.74 	&29.19 	&23.66 	&15.52 	&79.61 	&45.09 	&94.12 	&49.16 	&76.54\\
    \midrule
    \multirow{3}{*}{MSDA} & LAB &54.39 &80.20 	&49.29 	&87.36 	&\textbf{39.60} 	&42.78 	&30.01 	&17.58 	&18.94 	&78.35 	&39.29 	&93.58 	&50.66 	&79.45\\
    ~ & MS2PL & 53.26 & 78.52 	&52.66 	&87.39 	&30.77 	&39.91 	&30.41 	&23.45 	&15.53 	&76.37 	&36.43 	&93.37 	&50.27 	&77.34\\
    ~ & Ours & \textbf{57.16} 		&80.35 	&53.24  	&\textbf{87.93} 	&32.46  	&\textbf{48.03}  	&30.97  	&27.47  	&\textbf{19.32} 	&\textbf{80.40} 	&\textbf{50.06} 	&\textbf{94.34} 	&\textbf{56.31} 	&\textbf{82.18} \\ 
    \bottomrule
  \end{tabular}
  }
\end{table*}

\begin{table*}
  \caption{Comparison with the state-of-the-art DA methods on indoor scenarios. * indicates the results provided in other papers~\citep{zhang2024behind}.}
  \label{indoor_table}
  \centering
  \setlength{\tabcolsep}{6pt} 
  \small
  \begin{tabular}{c|c|c|llllllll}
    \toprule
Standards&Methods&mIoU&Ceiling&Chair&Door&Floor&Sofa&Table&Wall&Window\\
    \midrule
    \multirow{2}{*}{Source-only}& Single-best*  &63.73  &\textbf{90.63} 	&62.30 	&24.79 	&92.62 	&35.73 	&73.16 	&78.74 	&51.78 \\
    ~& Combined &65.97  &89.79 	&64.19 	&25.09 	&94.00 	&42.28 	&\textbf{77.34} 	&78.76 	&56.28 \\
    \midrule
    \multirow{4}{*}{Single-best DA}& DACS & 64.11   &87.97 	&59.50 	&37.52 	&92.50 	&30.67 	&71.14 	&79.20 	&54.35  \\
    ~& MIC &63.71   &88.58 	&59.97 	&24.19 	&92.08 	&42.73 	&68.19 	&77.88 	&56.06  \\
    ~& PiPa &63.86   &89.11 	&61.45 	&24.58 	&92.85 	&38.42 	&71.79 	&78.13 	&54.55  \\
    ~& Trans4PASS+* &67.16   &90.04 	&64.04 	&42.89 	&91.74 	&38.34 	&71.45 	&81.24 	&57.54  \\
    \midrule
    \multirow{4}{*}{Combined DA}&DACS & 65.34  &89.56 	&63.43 	&15.87 	&94.07 	&44.38 	&76.98 	&77.54 	&60.87 \\
    ~& MIC &67.31  &89.49 	&64.30 	&29.76 	&93.81 	&42.05 	&76.37 	&79.17 	&63.53 \\
    ~& PiPa &65.58  &89.60 	&63.44 	&21.29 	&94.07 	&35.62 	&76.92 	&78.30 	&65.44 \\
    ~& Trans4PASS+ &66.65   &89.37 	&64.29 	&14.95 	&94.03 	&\textbf{49.05} 	&76.00 	&77.53 	&\textbf{68.01}   \\
    \midrule
    \multirow{3}{*}{MSDA} & LAB &65.65  &89.74 	&61.32 	&28.54 	&93.71 	&37.85 	&75.20 	&79.29 	&59.52 \\
    ~ & MS2PL & 68.10 & 89.74 	&63.01 	&44.68 	&92.94 	&35.08 	&72.13 	&80.93 	&66.25\\
    ~ & Ours & \textbf{70.29}  &89.88  	&\textbf{66.02}  	&\textbf{48.68}  	&\textbf{94.36} 	&47.22  	&76.07  	&\textbf{82.65}  	&57.41   \\ 
    \bottomrule
  \end{tabular}
\end{table*}

\textbf{Training Details.}
For both outdoor and indoor settings, we use a single NVIDIA A100 GPU with 40 GB of memory to conduct the experiments.
First, we use the source domain samples to obtain the source pre-trained model, excluding target domain samples and without applying the deformation transformation. Specifically, we train the pinhole branch $h_{pin}$ and panoramic branch $h_{pan}$ using samples from source domains $\{U^i\}^{M}_{i=1}$ and $\{V^i\}^{N}_{i=1}$ with the standard cross-entropy ($CE$) loss, respectively. Additionally, we train the target branch $h_{t}$ using the class-mixed images between source domains with the standard cross-entropy loss. We train 200 epochs for source pre-training. This procedure requires approximately 8 hours.
Second, we train for 40,000 iterations during the DTA4PASS multi-source domain adaptation procedure using target samples and the deformation transformation. This procedure takes about 36 hours.

For source pre-training, we use a batch size of 4, whereas for domain adaptation, we use a batch size of 1. The learning rate and weight decay are set to 2.5e-6 and 5e-5, respectively, for both the diffeomorphic deformation network $F$ and the discriminators $D$. They are optimized by Adam and RMSprop respectively. For the segmentation network, the learning rate is 5e-6 and the weight decay is 5e-4, using SGD as an optimizer.
Following \citet{jang2022dada}, during the training of discriminators $D$, we apply one-sided label smoothing and set the labels of the panoramic-like images to 0.9 instead of 1.0 to prevent overconfidence of the discriminators.

In DTA4PASS, the loss weight $\alpha$  is set to 20, and $\beta$ is set to 0.5 for outdoor and 1.0 for indoor, respectively. EMA rate $\gamma$ is set to 0.999 and pseudo label threshold $\eta$ is 0.95.

\subsection{Comparisons with State-of-the-art Methods}
For outdoor and indoor settings, the results are shown in Table~\ref{outdoor_table} and Table~\ref{indoor_table}, respectively. We have the following analysis:

(1) Combining the source domains to train the segmentation model results in better performance than using only the source pinhole domain or the source panoramic domain (52.39\% vs. 51.48\% for outdoor and 65.97\% vs. 63.73\% for indoor). This confirms the rationale behind our motivation for proposing the MSDA4PASS task, which utilizes real pinhole images and synthetic panoramic images to enhance the segmentation network's ability to perceive both real scenes and panoramic structures simultaneously.

(2) All DA methods surpass the baseline method Source-only, which only adopts the source domain pinhole and panoramic images without using the target domain panoramas. This indicates that, in the MSDA4PASS task, there is a significant domain gap between the source domains and the target domain, and careful consideration is required to bridge this gap in order to achieve better panoramic generalization.
However, directly combining the source domains does not lead to stable improvements in performance in some DA methods, \textit{e.g.}, Trans4PASS+ Single-best 55.24\% vs. Combined 54.50\% in outdoor. This is because the domain gap between source domains is not accounted for, especially in the MSDA4PASS task, where the source domains contain both pinhole and panoramic images with substantial differences in appearance.

(3) The MSDA method generally outperforms the DA method, particularly under the Combined DA setting. For example, in indoor scenarios, LAB outperforms DACS, MIC, and PiPa; in outdoor scenarios, MS2PL achieves performance second only to our proposed DTA4PASS method. This suggests that in the MSDA4PASS task, if the domain gap of the source domain is properly addressed, \textit{i.e.,} distortion and texture gap, the generalization performance of the segmentation network on the target domain can be further improved. This is one of the motivations behind our proposal of the DTA4PASS method. DTA4PASS first considers converting all pinhole images in the source domains into distorted images to reduce the distortion gap between pinhole and panoramic images. DTA4PASS further considers aligning the source and target domains to alleviate the texture gap and enhance the model's adaptability to real-world panoramic scenes.

(4) Our method outperforms previous state-of-the-art pinhole DA methods, panoramic DA methods, and pinhole multi-source DA methods by 3.62\%, 1.92\%, and 2.77\% in outdoor scenes and 2.98\%, 3.13\%, and 2.19\% in indoor scenes, respectively. 
Our method also achieves the best performance in outdoor categories such as Traffic sign (Tr.sign), Person, and Car, which are crucial for autonomous driving. In indoor scenes, our method achieves the best performance in the Door, Floor, and Wall categories, which is also beneficial for indoor scene perception.
These superior performances result from our proposed USM, which bridges the distortion gap between pinhole and panoramic images, as well as DGA, which alleviates the texture gap between the source and target domains.

\begin{figure}[!t]
\centering
\includegraphics[width=0.98\linewidth]{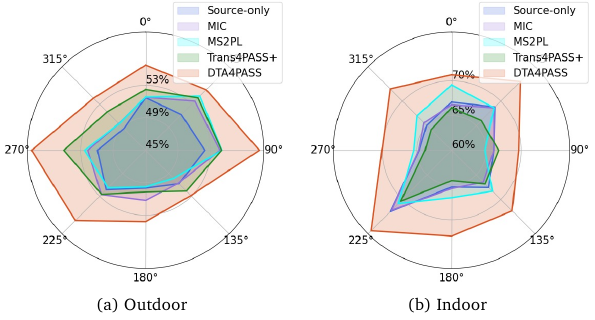}
\caption{Results of omnidirectional semantic segmentation on outdoor and indoor scenes. With the exception of the multi-source DA methods DTA4PASS and MS2PL, all other methods use a Combined DA setting.}
\label{fig:lidar}
\end{figure}

We further conduct experiments on omnidirectional semantic segmentation by dividing the panoramic images into 8 angles ($45^\circ$ for each) for separate testing, as illustrated in Figure~\ref{fig:lidar}. The results indicate that our method performs better than previous DA methods in all directions, whether indoors or outdoors. This demonstrates that our method achieves the best results in both global ($360^\circ$ in the panoramic scene) and local ($45^\circ$ in the omnidirectional scene) perception of real panoramic scenes. This improvement is due to our proposed DvD module providing feedback to the USM module through image-level and pixel-level discrimination, enabling the USM module to generate better deformations, thus enhancing distortion perception for the segmentation network. Moreover, the proposed DGA maintains the performance of the segmentation network in small field of view scenes ($45^\circ$ in omnidirectional scene) by aligning the source and target domains through the gating module in a pinhole-panoramic dual branch structure, where pin-like features are not ignored during training and inference.

\subsection{Ablation Study}
\begin{table}[!t]
  \caption{Module ablation for DTA4PASS. $h_t$ refers to the target branch, while UE indicates the uncertainty estimation.}
  \label{ablation_table_seg}
  \centering
  \setlength{\tabcolsep}{6pt} 
  \small
   \begin{tabular}{p{0.03\textwidth}p{0.03\textwidth}p{0.03\textwidth}p{0.03\textwidth}|cc}
    \toprule
    \multicolumn{4}{c}{Module} & \multicolumn{2}{c}{mIoU}\\
    $h_t$&UE&$\mathcal{L}_{adv}^{img}$&$\mathcal{L}_{adv}^{pix}$&Outdoor&Indoor \\
    \midrule
    ~&~&~&~& 54.25 	&67.53 \\
    \checkmark& ~&~&~& 55.74   	&67.89   \\
    \checkmark& \checkmark&~&~& 56.03   	&68.18   \\
    \checkmark& \checkmark&\checkmark&~& 56.78  	&69.93  \\
    \checkmark& \checkmark&\checkmark&\checkmark& \textbf{57.16} 	&\textbf{70.29}  \\

    \bottomrule
  \end{tabular}
\end{table}

\begin{table}[!t]
  \caption{\textcolor{mark}{Ablation study for the structure of gating module $g$.}}
  \label{ablation_table_gating}
  \centering
  \setlength{\tabcolsep}{5pt} 
  \small
   \begin{tabular}{c|ccccc}
    \toprule
    Structure & Add & Concat & MLP & Attention & Conv \\
    \midrule
   Outdoor& 56.79    & 56.12   & 57.03  & \textbf{57.19}    & 57.16  \\
   Indoor& 69.37    & 69.20   & 70.25   & \textbf{70.60}    & 70.29  \\
    \bottomrule
  \end{tabular}
\end{table}

\begin{table}[!t]
  \caption{\textcolor{mark}{Ablation study for the discriminator $D$.}}
  \label{ablation_table_dis}
  \centering
  \setlength{\tabcolsep}{3pt} 
  \small
   \begin{tabular}{c|ccccc}
    \toprule
    $D$ & PixelGAN & PatchGAN & StyleGAN2 & DvD (ours) \\
    \midrule
   Outdoor& 56.37    & 56.28   &56.62  & \textbf{57.16}  \\
   Indoor& 68.98    & 69.03   &69.51  & \textbf{70.29}  \\
    \bottomrule
  \end{tabular}
\end{table}

\begin{table}[!t]
  \caption{Sensitivity analysis of loss weight hyper-parameter $\alpha$.}
  \label{ablation_table_alpha}
  \centering
  \setlength{\tabcolsep}{6pt} 
  \small
   \begin{tabular}{c|ccccc}
    \toprule
    Scenario & $\alpha$=15 & $\alpha$=17.5 & $\alpha$=20 & $\alpha$=22.5 & $\alpha$=25 \\
    \midrule
   Outdoor& 57.09   & 57.11   & \textbf{57.16} & 57.14   & 57.14   \\
   Indoor& 70.20   & 70.24  & \textbf{70.29}  & 70.12   & 70.09  \\
    \bottomrule
  \end{tabular}
\end{table}

\begin{table}[!t]
  \caption{Sensitivity analysis of loss weight hyper-parameter $\beta$.}
  \label{ablation_table_beta}
  \centering
  \setlength{\tabcolsep}{6pt} 
  \small
   \begin{tabular}{c|ccccc}
    \toprule
    Scenario & $\beta$=0.1 & $\beta$=0.3 & $\beta$=0.5 & $\beta$=0.7 & $\beta$=1.0 \\
    \midrule
   Outdoor& 57.04  & \textbf{57.17} & 57.16 & 57.04  & 56.97  \\
   Indoor& 69.70  &69.86  &70.00  &70.17 &\textbf{70.29}  \\
    \bottomrule
  \end{tabular}
\end{table}

We perform detailed ablation experiments on the USM and DGA modules proposed in DTA4PASS, evaluating their effectiveness in both indoor and outdoor scenarios, as shown in Table~\ref{ablation_table_seg}. The following conclusions can be drawn:

(1) All the primary modules in DTA4PASS are effective. Performance progressively improves with the addition of each module. When all modules are removed (first row in Table~\ref{ablation_table_seg}), the baseline method refers to a configuration that includes only the pinhole branch $h_{pin}$ and panoramic branch $h_{pan}$, without applying any deformation to the source domain samples, while keeping all other settings identical to those of DTA4PASS. Although the baseline exhibits lower performance, it still outperforms several DA methods, such as DATR (54.05\% in outdoor) and Trans4PASS+ (67.16\% in indoor). This is because the source domain pinhole and panoramic images are processed separately, preventing feature confusion that arises from using the same head to segment pinhole and panoramic images with significant appearance differences.

(2) When the target branch $h_t$ is added to the proposed DGA module, performance improves further by 1.49\% and 0.36\% in outdoor and indoor settings, respectively. This is because the gating module $g$ in the target branch $h_t$ can dynamically assign pin-like features from the pinhole branch $h_{pin}$ and pan-like features from the panoramic branch $h_{pan}$ to both the source domain class-mixed images and the target domain panoramic images. Fully utilizing these two sets of features to perform segmentation on panoramic images results in better performance.

(3) After introducing uncertainty estimation between pin-like and pan-like features into the gating module $g$ of the proposed DGA, performance continues to improve. This is because uncertainty estimation reduces the differences between pin-like and pan-like features in the latent space. This helps the gating module $g$ fuse these two features and obtain consistent features, which is advantageous for the panoramic segmentation of the target branch $h_t$.

(4) When the proposed USM module is introduced with only the image-level adversarial loss $\mathcal{L}_{adv}^{img}$, the performance of the segmentation model improves by 0.75\% and 1.75\% in outdoor and indoor environments, respectively. This demonstrates the effectiveness of the USM module in converting all source pinhole images into distorted images, thereby reducing the appearance distortion gap between pinhole and panoramic images.

(5) After adding pixel-level adversarial loss $\mathcal{L}_{adv}^{pix}$ to the USM, the complete DTA4PASS proposed in this paper is obtained, and its performance improves further. This is because pixel-level adversarial loss can provide more local feedback to the deformation network $F$ to generate finer deformations, which is beneficial for the segmentation network $S$ to improve its perception of local deformations.

\begin{figure*}[!t]
\centering
\includegraphics[width=0.95\linewidth]
{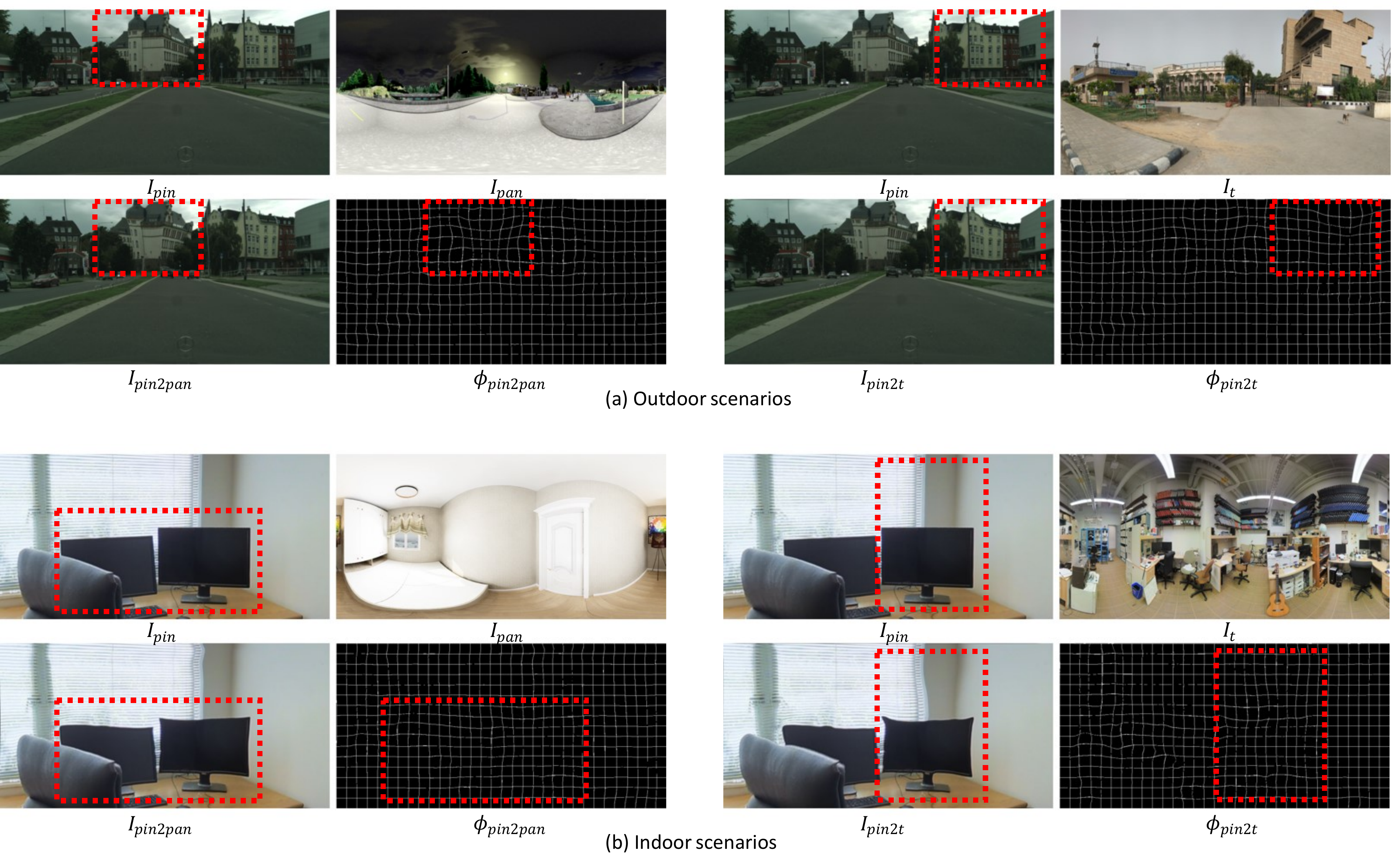}
\caption{Visualization of deformations. Given various reference images, the USM module can generate different deformation fields to transform the pinhole image into the corresponding deformed image.}
\label{fig:visualize_morph}
\end{figure*}

\begin{figure}[!t]
\centering
\includegraphics[width=0.98\linewidth]{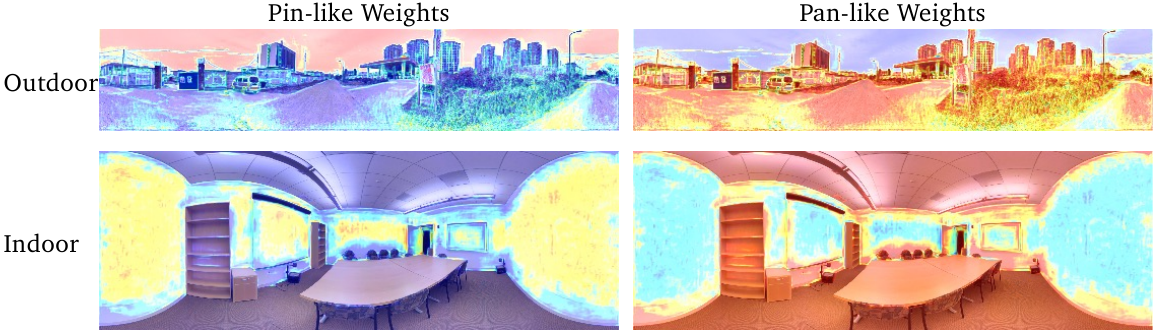}
\caption{Visualization of the gating weights in DGA. The redder the region, the higher the weight; the lighter the purple, the lower the weight.}
\label{fig:visualize_moe}
\end{figure}

\textcolor{mark}{
Meanwhile, we conduct ablation studies on the structure of the gating module $g$ in the DGA module, as shown in Table~\ref{ablation_table_gating}. The convolutional networks used in the experiments are replaced with addition, concatenation, MLP, and Attention~\cite{vaswani2017attention}, respectively. The following conclusions can be drawn: (1) Without using a gating module, \textit{i.e.}, without allocating pin-like and pan-like features, directly adding or concatenating these features for segmentation leads to a performance drop. This is because the degree of distortion varies across different regions of the panoramic image, and using pin-like and pan-like features with different weights for segmentation is more advantageous. This demonstrates the effectiveness of the proposed gating module. (2) Replacing convolutional networks with other architectures, such as MLP and Attention, may result in slight performance changes, but still outperforms those without using gating module. Using Attention as the gating module slightly improves performance, and more complex and refined designs may further enhance performance.
}

\textcolor{mark}{
We further conduct ablation studies on the effectiveness of the proposed DvD, as shown in Table~\ref{ablation_table_dis}. We replace DvD with pixel-level PixelGAN~\cite{isola2017image}, patch-level PatchGAN~\cite{isola2017image}, and image-level StyleGAN2~\cite{karras2020analyzing} discriminators, respectively. The experimental results indicate that the proposed DvD is effective and outperforms the other discriminators. This is because DvD simultaneously performs both image-level and pixel-level discrimination on the input image. This enables the deformation network to more effectively perceive both global and local distortion differences between pinhole and panoramic images, thereby generating improved deformation fields and enhancing performance.
}

As for the sensitivity analysis of the loss weight hyperparameters $\alpha$ for the deformation network $F$ in Eq.~(\ref{eq:loss_morph}) and $\beta$ for the segmentation network $S$ in Eq.~(\ref{eq:loss_seg}), they are presented in Table~\ref{ablation_table_alpha} and Table~\ref{ablation_table_beta}, respectively. DTA4PASS is relatively insensitive to both loss hyperparameters, demonstrating robustness.

\subsection{Qualitative Analysis}

\textbf{Deformation Visualization of USM.} We visualize the deformation of pinhole images in both outdoor and indoor scenes, as shown in Figure~\ref{fig:visualize_morph} (a) and (b). For the same pinhole image, different panoramic images cause the deformation network $F$ to generate different deformation fields, transforming the pinhole image into corresponding distorted images.
The deformed image exhibits distortion, and providing such images to the segmentation network enhances the network's ability to perceive distortion.
Moreover, USM can generate subtle deformations, thanks to the proposed DvD, which provides adversarial feedback at both the image and pixel levels to the deformation network $F$.

\textbf{Gating Weight Visualization of DGA.} We visualize the selected weights of pin-like and pan-like features for target domain panoramas using the gating module $g$ in DGA, as shown in Figure~\ref{fig:visualize_moe}. On the one hand, DGA tends to allocate more pan-like features to the panoramic images of the target domain, especially in large distortion areas. This is beneficial because these areas contain significant panoramic structures, making them more suitable for assigning pan-like features. 
On the other hand, DGA assigns fewer pin-like features, found only in small portions of areas where deformation distortion is not severe, such as Sky and Wall. Overall, these visualization results demonstrate the rationale and effectiveness of the pin-like and pan-like feature allocation by DGA. Moreover, the working mechanism of the gating module can be regarded as an attention mechanism, because the boundaries of pin-like and pan-like regions are obvious and usually located at object boundaries, which is beneficial for the segmentation network to accurately identify object boundaries.

\begin{figure*}[!t]
\centering
\includegraphics[width=0.98\linewidth]{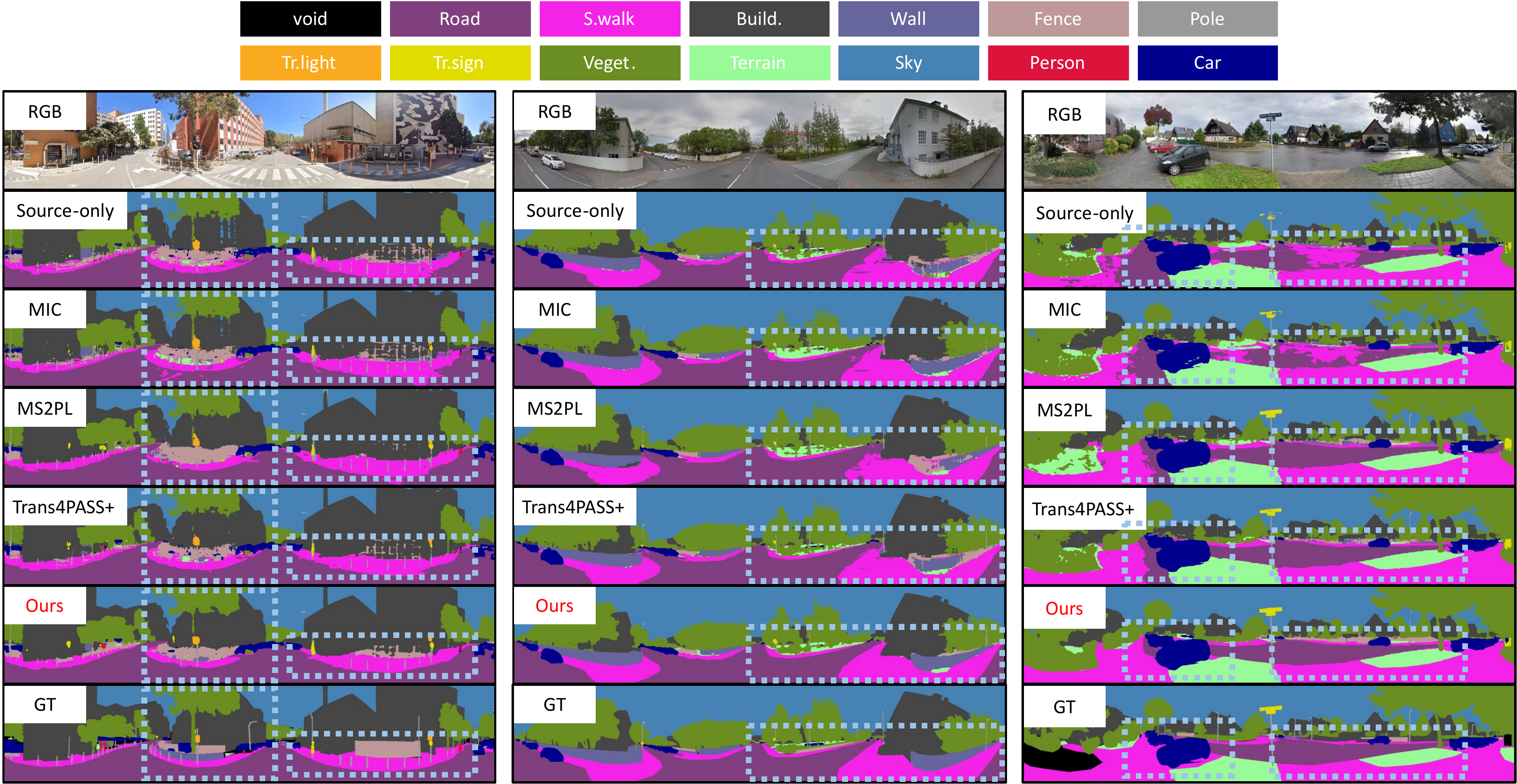}
\caption{\textcolor{mark}{Visualizations of panoramic semantic segmentation in outdoor settings. Please zoom in for a better view.}}
\label{fig:outdoor_seg}
\end{figure*}

\begin{figure}[!t]
\centering
\includegraphics[width=0.98\linewidth]{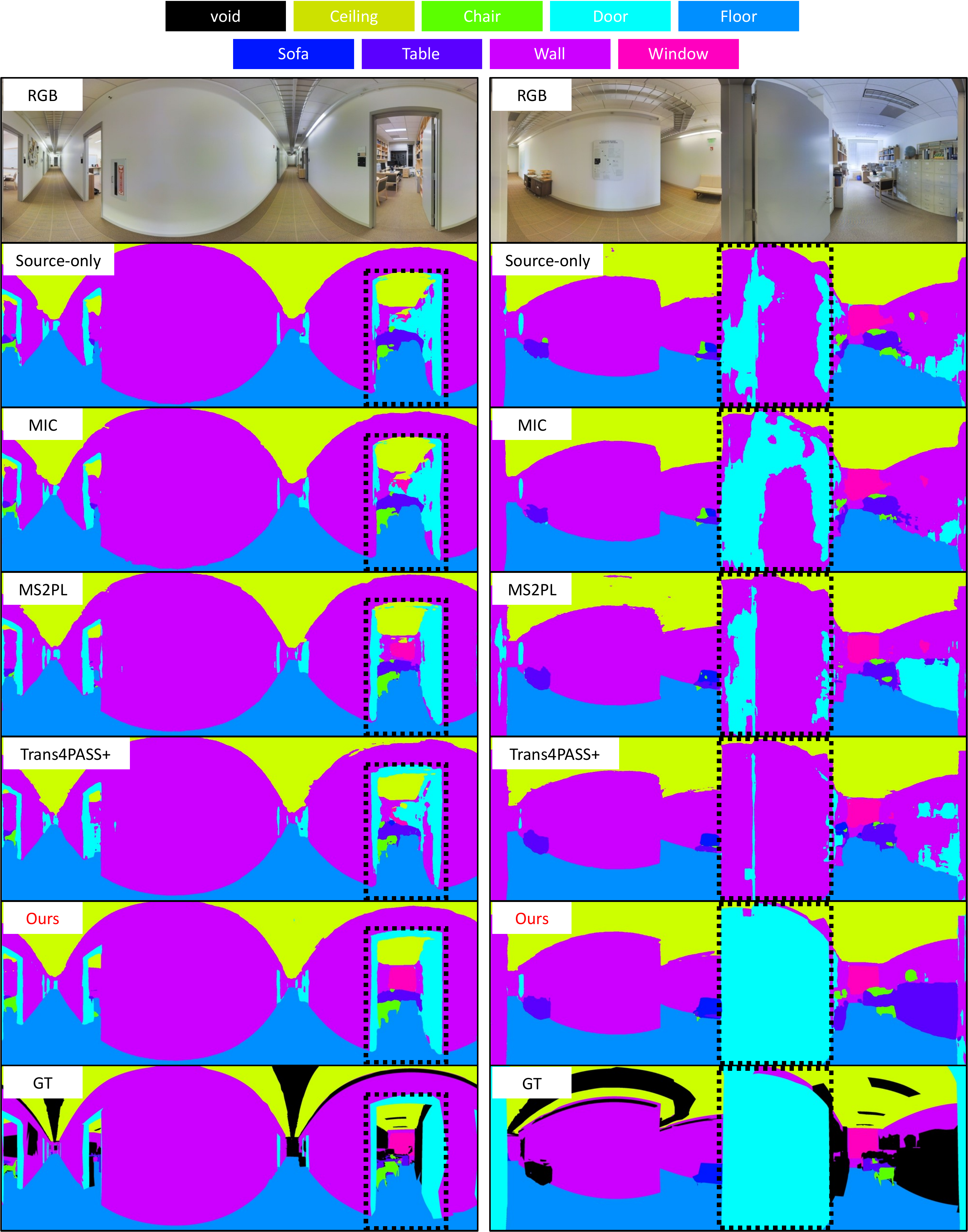}
\caption{Visualizations of panoramic semantic segmentation in indoor settings. Please zoom in for a better view.}
\label{fig:indoor_seg}
\end{figure}

\begin{figure}[!t]
\centering
\includegraphics[width=0.98\linewidth]
{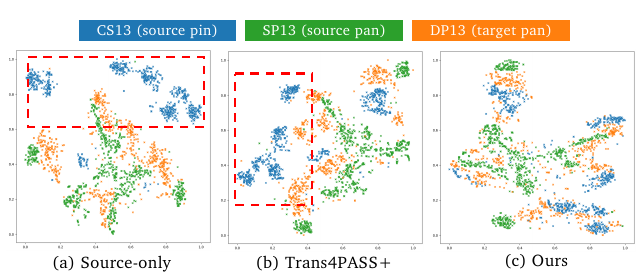}
\caption{t-SNE visualizations before and after domain adaptation by different methods in outdoor scenes.}
\label{fig:outdoor_tsne_small}
\end{figure}

\textbf{Segmentation Results.} We present the visualization results of DTA4PASS and other DA methods for panoramic semantic segmentation on outdoor settings in Figure~\ref{fig:outdoor_seg}. 
It can be observed that our method is more accurate in recognizing the boundaries of curved regions compared to other methods (the left part of the results in Figure~\ref{fig:outdoor_seg}), thanks to the fact that samples we provide to the segmentation network $S$ are source distorted images transformed through USM and panoramic images.
Moreover, our method is more accurate than others in identifying large-scale regions, and there is no occurrence of category mutation in intermediate regions (the middle and right parts of the results in Figure~\ref{fig:outdoor_seg}). This is because our DGA extracts pin-like and pan-like features separately, and then reduces the differences through the gating and uncertainty estimation modules, enabling the segmentation network to extract more consistent features from the target domain panoramas, thereby avoiding local mutations.

We further visualize the indoor segmentation results in Figure~\ref{fig:indoor_seg}. It is shown that our DTA4PASS not only accurately identifies curved areas in panoramic-like areas (on the right side of Figure~\ref{fig:indoor_seg}), but also accurately identifies areas in pinhole-like areas (on the left side of Figure~\ref{fig:indoor_seg}). This demonstrates the effectiveness of our proposed USM, which transforms source pinhole images into distorted images, and DGA, which allocates pixel-level pin- and pan-like features to the target panoramic images.

\textbf{t-SNE Visualization.} We further visualize the multi-source domain adaptation results using t-SNE, as shown in Figure~\ref{fig:outdoor_tsne_small}. Without any domain adaptation (Source-only), the gap between the three domains is relatively large. After the combined domain adaptation with panoramic DA method Trans4PASS+, the domain gap is reduced, but it is still easy to distinguish between pinhole (CS13) and panoramic images (SP13, DP13). After domain adaptation with our DTA4PASS, the three domains overlap with each other, and the gap is greatly reduced, thanks to the transformation of pinhole images into distorted images by USM and the alignment of pin- and pan-like features by DGA.

\section{Discussions and Limitations}
\label{sec:dis}

\begin{figure}[!t]
\centering
\includegraphics[width=0.98\linewidth]{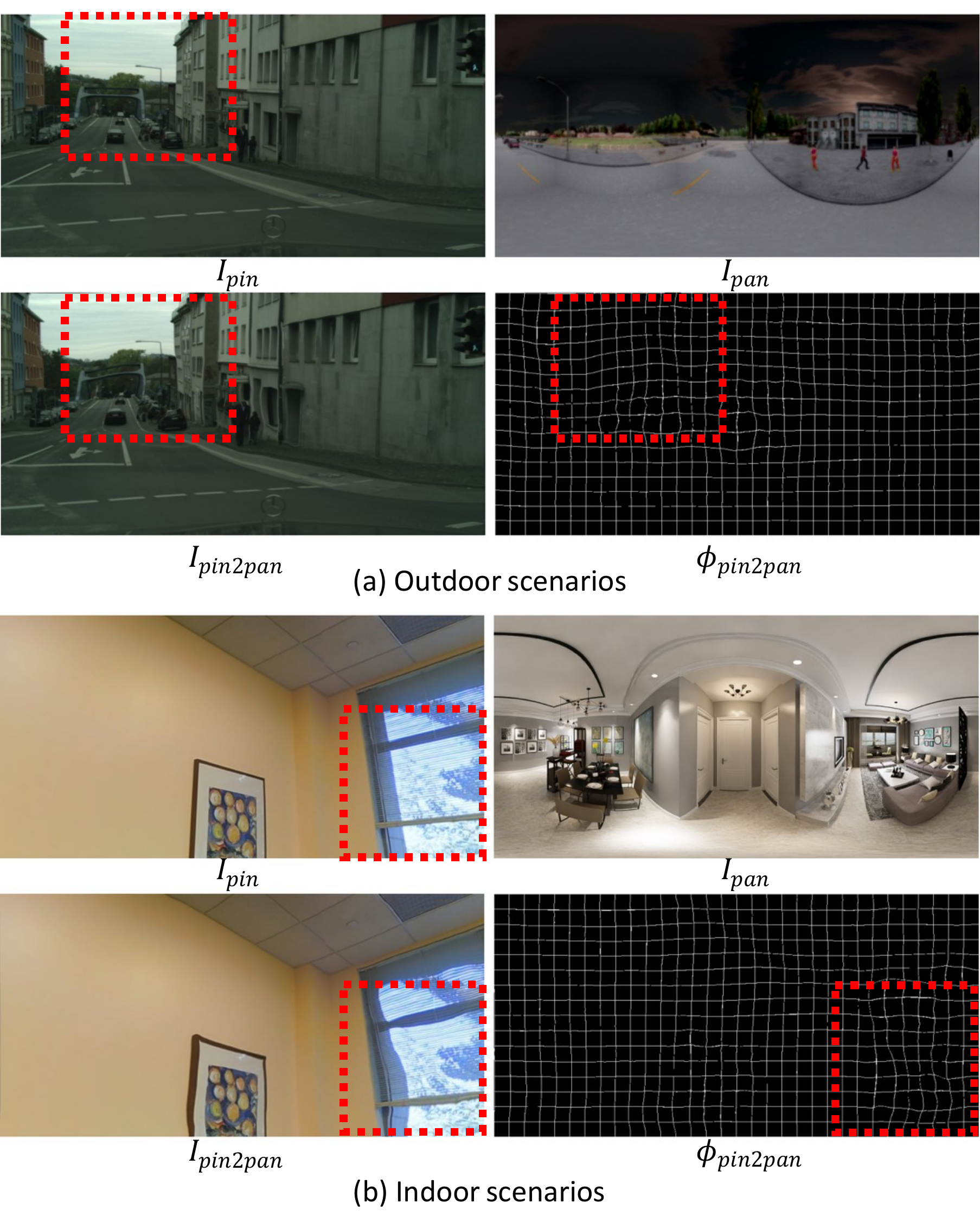}
\caption{\textcolor{mark}{Failure cases of USM.}}
\label{fig:visualize_morph_bad}
\end{figure}

\begin{figure*}[!t]
\centering
\includegraphics[width=0.98\linewidth]{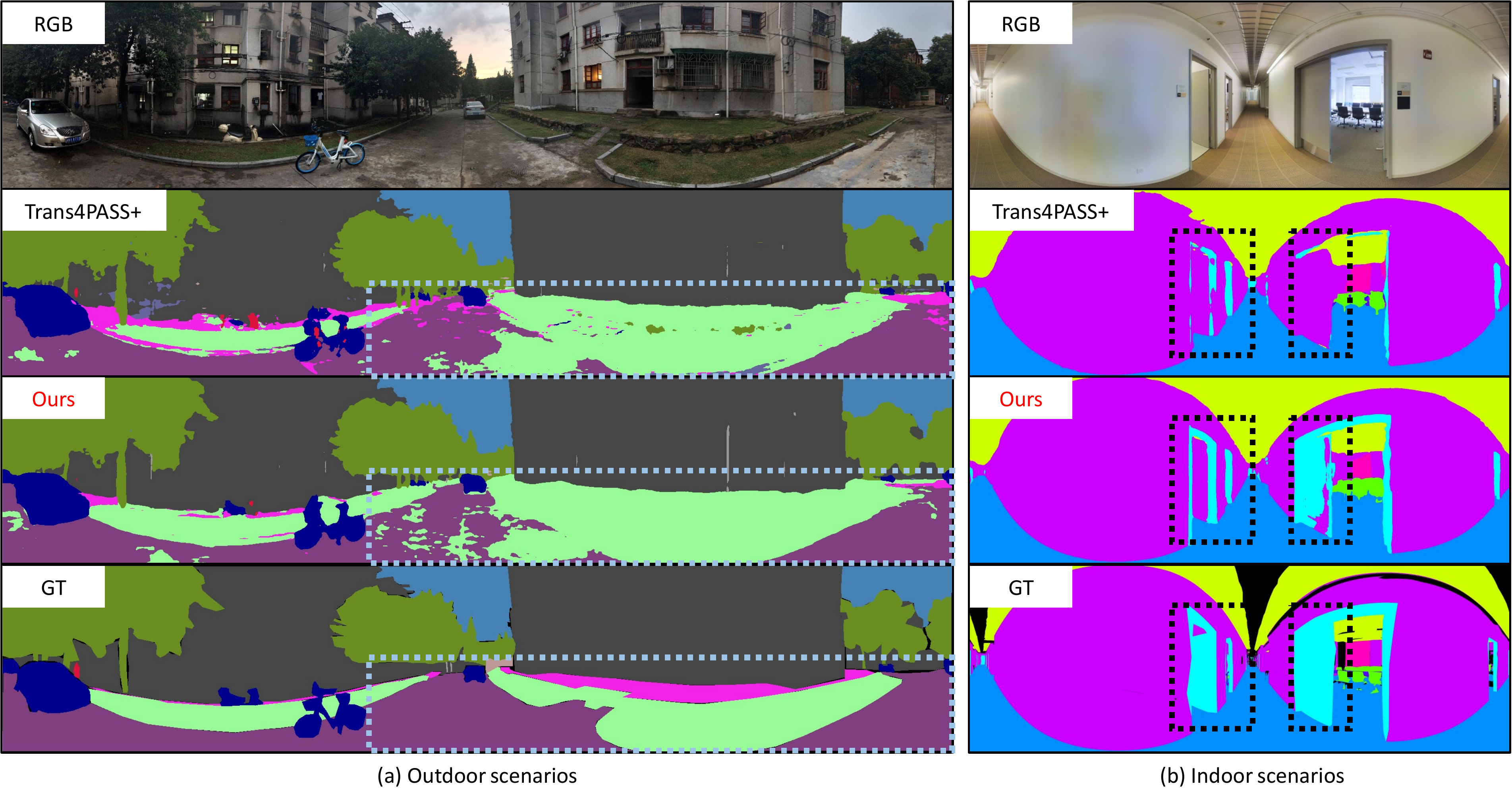}
\caption{\textcolor{mark}{Failure cases of the proposed DTA4PASS.}}
\label{fig:seg_bad}
\end{figure*}

We introduce a new task, named MSDA4PASS, for the panoramic community. It is both novel and practical because it enables flexible knowledge transfer from various distorted forms of samples in the source domains to the target panoramic domain. The proposed DTA4PASS method is also easy to promote, as the USM module can be extended to other panoramic tasks, such as panoramic classification, detection, and panoptic segmentation. Due to the limited availability of datasets, we have considered only two source domains for pinhole and panoramic images. However, our method can be expanded to include more source domains, which would further enhance generalization.

Although the deformation results produced by the USM module are promising, USM also has certain limitations. Training the USM incurs additional costs, as it is not needed during the inference phase of the segmentation network. \textcolor{mark}{Furthermore, the unpaired adversarial training method used by USM can yield unsatisfactory results, as it is unable to fully simulate the distorted structure of panoramic images, providing only partial simulations, as shown in Figure~\ref{fig:visualize_morph_bad}~(a). This is due to unpaired unsupervised training, which causes confusion in the deformation network regarding which regions need to be deformed. Introducing supervised paired pinhole and distorted images could greatly alleviate this issue. The deformation field may not be sufficiently smooth, as shown in Figure~\ref{fig:visualize_morph_bad}~(b). Introducing smoothing constraints, such as Gaussian smoothing, may improve the deformation field, which is part of our future work. Moreover, using more powerful and complex adversarial losses to train the deformation network, such as CycleGAN~\cite{zhu2017unpaired,zhao2021madan,wu2024stegogan} and LSGAN~\cite{mao2017least,tasar2020standardgan,chen2020domain} loss, may improve the quality of the generated deformation field.} Making USM more efficient and achieving better deformation transformations is among our future work.

\textcolor{mark}{
While the proposed DTA4PASS outperforms existing state-of-the-art (SOTA) methods and achieves good segmentation results, there are still some failure cases, as shown in Figure~\ref{fig:seg_bad}. The current domain adaptation methods, including our method and the previous SOTA method, Trans4PASS+, are not sufficient for panoramic boundary segmentation, as shown on the right side of Figure~\ref{fig:seg_bad} (a) and (b). Utilizing the powerful zero-shot segmentation capability of SAM~\cite{kirillov2023segment} can alleviate this problem, and it is part of our future work.
Meanwhile, the segmentation model overly relies on pseudo-labels, resulting in unsatisfactory segmentation of uncommon object shapes. As shown on the left side of Figure~\ref{fig:seg_bad} (b), the `Door' should encompass both the door frame and the door panel, but DTA4PASS and Trans4PASS+ fail to segment the opened door panel correctly. This is because these methods are all pseudo-label-based, which may lead to overfitting. A better pseudo-label refinement method could alleviate this issue.
}

\section{Conclusion}
\label{sec:conclusion}

In this paper, we propose a novel task, Multi-source Domain Adaptation for Panoramic Semantic Segmentation (MSDA4PASS), with the goal of leveraging a large amount of labeled real pinhole images and low-cost synthetic panoramic images to enhance the perception of real panoramic scenes. To address the two main challenges in the MSDA4PASS task: the distortion gap between the pinhole and panoramic images and the texture gap between the source and target images, we propose Deformation Transform Aligner for Panoramic Semantic Segmentation (DTA4PASS), which consists of two components, Unpaired Semantic Morphing (USM) and Distortion Gating Alignment (DGA). First, USM transforms all source pinhole images into distorted images in an adversarial manner to reduce the distortion gap. Second, DGA assigns pixel-level pin- and pan-like features to source class-mixed and target domain samples, incorporating an uncertainty estimation module to align the distorted source domains with the target domain, thus mitigating the texture gap. Extensive experiments conducted in both outdoor and indoor settings demonstrate that our method achieves state-of-the-art performance on the MSDA4PASS task.

\section*{Acknowledgements}
\label{sec:Acknowledgements}
This work is supported by the National Natural Science Foundation of China (No. 62441202) and CCF-DiDi GAIA Collaborative Research Funds.

\bibliographystyle{cas-model2-names}

\bibliography{cas-refs}
\end{document}